\documentclass[sn-mathphys-ay]{sn-jnl}

\usepackage{graphicx}%
\usepackage{multirow}%
\usepackage{amsmath,amssymb,amsfonts}%
\usepackage{amsthm}%
\usepackage{mathrsfs}%
\usepackage[title]{appendix}%
\usepackage{xcolor}%
\usepackage{colortbl}
\usepackage{textcomp}%
\usepackage{manyfoot}%
\usepackage{booktabs}%
\usepackage{algorithm}%
\usepackage{algorithmicx}%
\usepackage{algpseudocode}%
\usepackage{listings}%
\usepackage{subcaption}
\usepackage{tabularx}
\usepackage{xspace}
\usepackage{adjustbox}

\theoremstyle{thmstyleone}%
\theoremstyle{thmstyletwo}%

\theoremstyle{thmstylethree}%
\def\eg{\emph{e.g.}\xspace} 

\def\ie{\emph{i.e.}\xspace}

\begin{document}

\title[Article Title]{Towards Generalizing Temporal Action Segmentation to Unseen Views}


\author[1,2]{\fnm{Emad} \sur{Bahrami}}
\equalcont{These authors contributed equally to this work.}
    
\author[1,2]{\fnm{Olga} \sur{Zatsarynna}}
\equalcont{These authors contributed equally to this work.}

\author[3]{\fnm{Gianpiero} \sur{Francesca}}
\author[1,2]{\fnm{Juergen} \sur{Gall}}

\affil[1]{\orgdiv{University of Bonn}, \country{Germany}}

\affil[2]{\orgdiv{Lamarr Institute for Machine Learning and Artificial Intelligence}, \country{Germany}}

\affil[3]{\orgdiv{Toyota Motor Europe}, \country{Belgium}}


\abstract{While there has been substantial progress in temporal action segmentation, the challenge to generalize to unseen views remains unaddressed. Hence, we define a protocol for unseen view action segmentation where camera views for evaluating the model are unavailable during training. This includes changing from top-frontal views to a side view or even more challenging from exocentric to egocentric views. Furthermore, we present an approach for temporal action segmentation that tackles this challenge. 
Our approach leverages a shared representation at both the sequence and segment levels to reduce the impact of view differences during training. We achieve this by introducing a sequence loss and an action loss, which together facilitate consistent video and action representations across different views.
The evaluation on the Assembly101, IkeaASM, and EgoExoLearn datasets demonstrate significant improvements, with a $12.8\%$ increase in F1@50 for unseen exocentric views and a substantial $54\%$ improvement for unseen egocentric views. }






\maketitle

\section{Introduction}
\label{sec:intro}
Temporal action segmentation identifies action segments and their respective durations within a long untrimmed video. This task has numerous practical applications, including assistive technologies, production line monitoring, and animal behavior analysis. While there has been major progress in temporal action segmentation~\citep{xudon2022dtl, behrmann2022uvast, aziere2022multistage, ltc2023bahrami, liu2023diffusion, gong2023activity}, the generalization to unseen views has not been tackled. This poses additional challenges when such approaches are deployed in real environments. While it is possible to capture training data from multiple views~\citep{sener2022assembly101, Grauman2023EgoExo4DUS}, the camera setting used during data collection often does not cover the operation conditions since the environment and regulations often impose constraints where cameras can be mounted. Re-collecting training data for a new camera view is often infeasible or economically impractical. While knowledge distillation techniques \citep{quattrocchi2023synchronization} between two camera views can be used to minimize the annotation process for newly recorded data, it still requires recording new training data with the new and an old camera view. 
To remove the need for new data acquisition for unseen camera views, we propose a method that addresses these issues directly during training.                 

\begin{figure*}[tb]
    \centering
    \includegraphics[width=1.0\linewidth]{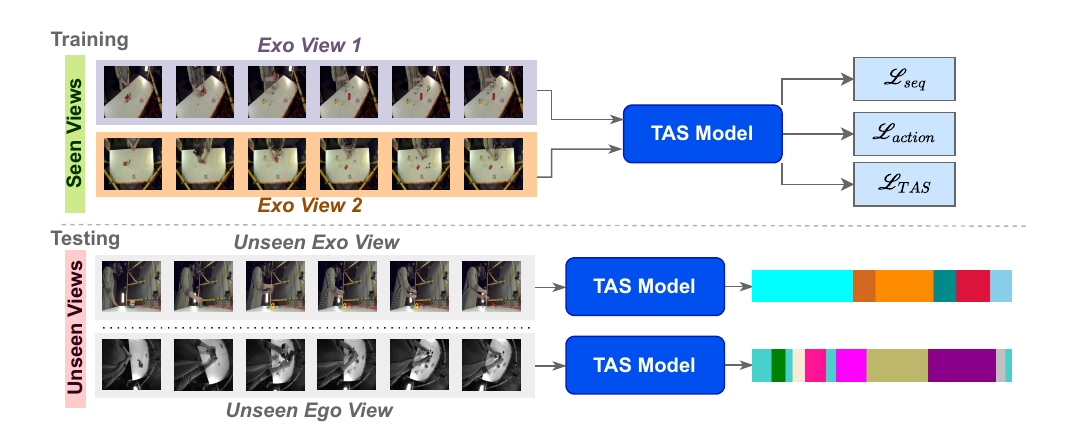}
    \caption{In this work, we address the problem of temporal action segmentation (TAS) on unseen views. During training, we observe at least two different views of long video sequences that have been frame-wise annotated by the occurring actions. Despite using a standard loss for temporal action segmentation $\mathcal{L}_{TAS}$, we propose a sequence $\mathcal{L}_{seq}$ and an action loss $\mathcal{L}_{action}$ that increase the generalization to unseen views without reducing the accuracy on seen views. The model can then be deployed in a setting with unseen views that are very different to the views in the training set.}
    \label{fig:teaser}
\end{figure*}

Figure~\ref{fig:teaser} shows the setting for temporal action segmentation on unseen views. Given annotated training data that includes at least two different views, the goal is to temporally segment the actions in a video from an unseen view. We focus in particular on scenarios where there is a large discrepancy between the views in the training data and the view in the test data. This includes the change from top-frontal views to a side view or even more challenging from exocentric views to an egocentric view. It is important to emphasize that the unseen views are not available during training and we aim to train a network for temporal action segmentation in a way such that it learns a temporal embedding where view information is as much as possible discarded. This is achieved by leveraging a Siamese network with weight sharing at two different temporal levels. While the proposed sequence loss aims to learn a video representation that is similar across views, the proposed action loss aims to learn an action representation that is similar across views. 

In our evaluation, we examine the model's generalization from exocentric views to both an unseen exocentric view and an unseen egocentric view. To evaluate our method, we establish an unseen view action segmentation setting for the Assembly101~\citep{sener2022assembly101} and IkeaASM~\citep{BenShabat2020TheIA} datasets. Furthermore, we evaluate our approach on the EgoExoLearn dataset~\citep{huang2024egoexolearn}. The experimental results show that the sequence and action loss complement each other and that the proposed approach increase the F1@50 by $12.8\%$ on unseen exocentric views and by $54\%$ on unseen egocentric views of Assembly101. Even for the seen views, there is a slight increase by $3\%$ since increasing generalization to unseen views also decreases overfitting.     

\section{Related Work}
\label{sec:rel_work}
\subsection{Fully-Supervised Temporal Action Segmentation}
Temporal action segmentation involves predicting dense classification scores for video frames. Initially, early approaches~\citep{rohrbach2012, karaman2014} relied on sliding-window non-maxima suppression. Additionally, temporal modeling of actions using Hidden Markov Models (HMMs) was explored by~\cite{kuehne2014language, kuehne2016end}, \cite{tang2012latent}, and \cite{richard2016temporal}.
Subsequent methods~\citep{lea2017temporal, farha2019ms,  huang2020improving} adopted the Temporal Convolution Network (TCN) for temporal action segmentation. \cite{lea2017temporal} proposed an encoder-decoder TCN network, while \cite{farha2019ms} introduced a multi-stage TCN (MS-TCN). Further methods were proposed for prediction refinement, \textit{i.e.} boundary regression network~\citep{ishikawa2021asrf} and graph convolution-based model~\citep{huang2020improving}.
Following the widespread adoption of the Transformer architecture~\citep{vaswani2017attention}, attention-based approaches~\citep{yi2021asformer, behrmann2022uvast, ltc2023bahrami, aziere2022multistage} emerged. Additionally, \cite{li2022bridge} combined vision transformers with the pre-trained vision-language model~\citep{wang2021actionclip} for feature extraction, while \cite{liu2023diffusion} utilized diffusion models~\citep{sohl2015NonEquTherm, song2019generative, ho2020denoisingDiff, song2021scorebased} for temporal action segmentation.
Another research direction focused on leveraging additional constraints to improve temporal action segmentation quality~\citep{xudon2022dtl, gong2023activity}, agnostic to the underlying architecture.

In contrast to these approaches, we focus on the task of unseen view temporal action segmentation, presenting challenges not previously addressed in research. While some prior works~\citep{Ghoddoosian2022WeaklySupervisedOA, quattrocchi2023synchronization} have integrated multiple views into their methodologies, our approach exhibits significant differences. In the work by \cite{Ghoddoosian2022WeaklySupervisedOA}, frame-wise correspondences between multiple views are exploited to generate pseudo-ground truth for weakly-supervised online action segmentation. 
\cite{quattrocchi2023synchronization} employ a knowledge distillation method for exo-to-ego model transfer, training an action segmentation model using a single exocentric view and subsequently adapting the model to an egocentric view.

Our approach specifically addresses a scenario where multiple views are available during the training of an action segmentation method. However, during the deployment of the method, the camera settings differ from the training setting. Therefore, the model needs to exhibit robustness to unseen views during testing. This setting is particularly relevant to real-world applications where a model is developed for a fixed camera setting, but the customers, where the system is deployed later, have a very different camera setting.

\subsection{View-invariant Action Recognition}
The task of view-invariant action recognition is to classify actions in videos captured from unseen viewpoints. Early studies~\citep{Imran2008CrossviewAR, Rao2002ViewInvariantRA, weinland2010robust, farhadi2008wrongviewpoint} in this domain attempted to manually design view-invariant video features that work well across all viewpoints. 
More recent work has focused on learning representations for this task.
Various modalities, like depth~\citep{Vyas2020MVAR, li2018unsupervised}, skeleton or motion~\citep{luo2017unsupervised, li2018unsupervised} were utilized by different works to achieve view-invariance of the resulting representations. In particular, skeleton-based methods have seen wide adaptation with many models proposed based on architectures like RNN~\citep{Sordoni2015AHR, shahroudy2016ntu, zhang2017view, liu2017skeleton}, TCN~\citep{Kim2017Interpretable3H}, GCN~\citep{Chen2021ChannelwiseTR, cheng2020shiftgcn, ke2022towards, Liu2020DisentanglingAU, shi2019skeleton, shi2019two, yan2018spatial, ye2020dynamic, chen2021MultiScaleST}, and more recently Transformer~\citep{plizzari2021skeleton, Duan_2023_ICCV, Gao2022FocalAG, shi2020decoupled, yang2021unik, Zhang2021STSTSS}.

A different line of work~\citep{mahasseni2013latent, Vyas2020MVAR, Shah_2023_WACV, Das2021ViewCLRLS, shah2024mv2mae, Piergiovanni2021RecognizingAI, Zheng2016CV, Liu2011KT} focuses on view-invariant action recognition solely from RGB data avoiding the usage of additional modalities. In recent years, due to the emergence of large-scale egocentric datasets~\citep{sener2022assembly101, Grauman2021Ego4DAT, Damen2022RESCALING, Grauman2023EgoExo4DUS, Sigurdsson2018ActorAO, Sigurdsson2018CharadesEgoAL}, a separate direction of ego-exo invariant action recognition has attracted increasing attention~\citep{Sermanet2017TimeContrastiveNS, Wang2023LearningFS, xue2023learning, Truong2023CrossviewAR, Yu2019WhatIS, Bilge2015AR}. 
Additionally, some methods for exo-to-ego knowledge transfer~\citep{ego-exo, Choi_2020_WACV, ARDESHIR201861} were introduced, allowing the utilization of third-view pretraining for egocentric action recognition. Following this direction, \citet{ohkawa2023exo2egodvc} has introduced the new task of cross-view (ego-to-exo) knowledge transfer for dense video captioning. 

\subsection{Siamese Networks}
Siamese networks~\citep{Bromley1993siamese} have been widely utilized in tasks such as face verification~\citep{taigman5closing}, one-shot image recognition~\citep{koch2015siamese}, tracking~\citep{bertinetto2016fully}, and representation learning~\citep{assran2022masked, chen2021simsiam}. Their ability to learn discriminative features by comparing pairs of inputs has also proven valuable in action recognition. Prior studies~\citep{sheng2019siamese, roy2018action, lu2024siammast, guanghui2024few} have explored their application in various action recognition settings, demonstrating their effectiveness in human activity recognition.

\begin{figure*}[tb]
    \centering
    \includegraphics[width=1.0\linewidth]{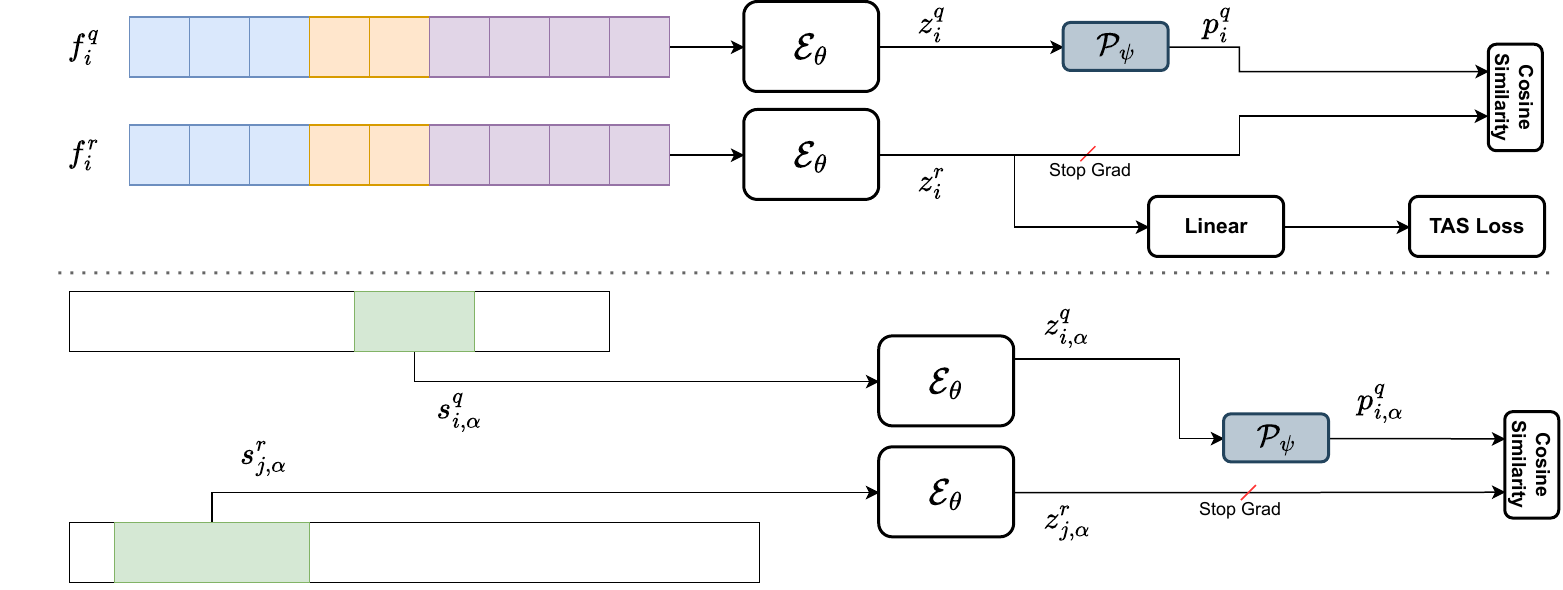}
    \caption{To train a network for temporal action segmentation that can be applied to views that are not part of the training data, we propose a sequence loss (top) and an action loss (bottom). The sequence loss takes two different camera views as input and computes the frame-wise similarity between the views (cosine similarity). In addition, a standard loss for temporal action segmentation (TAS) is applied to both views. The action loss takes randomly chosen action segments of the same action (green) as input and computes the frame-wise similarity between the action segments. The weights of $\mathcal{E}_{\theta}$ are shared between the two branches.}
    \label{fig:main_method_fig}
\end{figure*}

\section{Temporal Action Segmentation on Unseen Views}
\label{sec:method}

When approaches for temporal action segmentation are deployed in practice, the view often differs from the views that have been observed during training. In this work, we thus aim to increase the robustness of an approach for temporal action segmentation to unseen views. While we assume that the training sequences include at least two different views and have been annotated by the occurring actions, the test sequences contain a single view that has not been part of the training data. In our evaluation, we consider two cases: (a) generalization from exocentric views to an unseen exocentric view and (b) generalization from exocentric views to an unseen egocentric view. While the model should demonstrate robust performance on unseen views, it should maintain a consistent performance on the views encountered during training. 
It is important to emphasize that we do not assume any paired sequences between the seen and unseen views. Techniques like knowledge distillation can thus not be applied and we need to train a model that discards the view-specific information from the training data and focuses more on the inherent information of the actions independent of the view.   

To address these issues, we propose to learn a representation for temporal action segmentation by accounting for similarity between views. While similar embeddings for different views can be learned through contrastive learning approaches~\citep{chen2020simclr, he2020momentum, chopra2005learning}, they need large batch sizes or a memory bank, which is very costly for long video sequences as it is required for temporal action segmentation~\citep{ltc2023bahrami}. We thus build on Siamese networks~\citep{Bromley1993siamese, chen2021simsiam} with weight sharing and propose two loss terms that measure similarity between views at both the sequence and action levels. 

While Figure~\ref{fig:main_method_fig} illustrates the two loss terms, we discuss them in detail in Sections~\ref{sec:seq_loss} and \ref{sec:action_loss}. Section \ref{sec:task} first formalizes the task of temporal action segmentation on unseen views.   

\subsection{Problem Formulation}\label{sec:task}
During training, we have a set of N sequences $\{\mathcal{I}_{i}\}_{i=1}^{N}$, where each $\mathcal{I}_{i} = \{f_{i}^{k}\}_{k=1}^{M}$ represents the $i$th sequence recorded from $M$ distinct views denoted by $\mathcal{V}_{\text{seen}} = \{ {1}, \dots, {M} \}$. We assume that $M\geq2$. For each training sequence $\mathcal{I}_{i}$ with $T_i$ frames, the ground-truth actions, denoted by $c_{i} = [c_{1}, \dots, c_{T_i}]$, are given. 

For evaluation, videos from different views are provided, \ie, $\mathcal{V}_{\text{unseen}} \cap \mathcal{V}_{\text{seen}} = \emptyset$. Each video is represented by an $H$-dimensional input sequence $f \in \mathbb{R}^{T \times H}$ with $T$ frames where $f_{t}$ ($t\in\{1, \dots, T\}$) is a feature vector of size $H$. A network for temporal action segmentation $\mathcal{E}_{\theta}$ first encodes $f \in \mathbb{R}^{T \times H}$ into the representation $z \in \mathbb{R}^{T \times D}$. Then, a classification head is used to predict the frame-wise labels $\hat{c} = [\hat{c}_{1}, \dots, \hat{c}_{T}]$. The network for temporal action segmentation is commonly trained and tested on the same views $\mathcal{V}_{\text{seen}}$, and trained using an action segmentation loss, \eg, consisting of a frame-wise cross entropy and a smoothing mean squared error (MSE) loss~\citep{farha2019ms}. In this work, we propose two loss functions that increase the generalization to unseen views $\mathcal{V}_{\text{unseen}}$ without reducing the accuracy on seen views $\mathcal{V}_{\text{seen}}$.     
  
\subsection{Sequence Loss}
\label{sec:seq_loss}
To increase the robustness of a network for temporal action segmentation $\mathcal{E}_{\theta}$ to unseen views, we introduce two loss functions. The first loss term, which we term sequence loss, encourages the representations of all views of the same sequence to be similar. The loss is illustrated in Figure~\ref{fig:main_method_fig} (top).   

Given an input sequence $f_{i}^{q}$ from view $q$, we randomly sample $f_{i}^{r}$ from the same sequence $i$ but a different view $r \neq q$. 
Then, using $\mathcal{E}_{\theta}$, we obtain a pair of embeddings $z_{i}^{q}$ and $z_{i}^{r}$. While we add to $z_{i}^{r}$ a classification head to predict the frame-wise labels $\hat{c} = [\hat{c}_{1}, \dots, \hat{c}_{T}]$ and use a temporal action segmentation loss (TAS) for supervision, which will be described in Section~\ref{sec:joint_training}, a predictor head $\mathcal{P}_{\psi}$ maps $z_{i}^{q}$ to $p_{i}^{q} \in \mathbb{R}^{T \times D}$. The predictor head $\mathcal{P}_{\psi}$ is a 3 layer MLP with Gaussian Error Linear Units (GELU) activation with trainable parameters $\psi$. In the experiments, we will evaluate the impact of the predictor head, which improves the accuracy on unseen views.

For the sequence loss, we then compute the frame-wise cosine similarity between $(p_{i}^{q}, z_{i}^{r})$:
\begin{align}
    \mathcal{S}(p_{i}^{q}, z_{i}^{r}) = \sum_{t} \mathcal{D}(p_{i, t}^{q}, z_{i,t}^{r}), 
    \label{eq:seq_simlarity}
\end{align}
where $\mathcal{D}(\mathbf{a}, \mathbf{b}) = \frac{\mathbf{a} \cdot \mathbf{b}}{\|\mathbf{a}\| \|\mathbf{b}\|}$ is the cosine similarity. In the experiments, we show the benefit of a frame-wise similarity compared to pooling over the temporal dimension of the sequence.    

The overall sequence loss is finally computed in a symmetric way:
\begin{align}
    \mathcal{L}_{seq}= -\sum_{i} \frac{1}{2} \big[\mathcal{S}(p_{i}^{q}, z_{i}^{r}) + \mathcal{S}(p_{i}^{r}, z_{i}^{q}) \big].
    \label{eq:seq_loss}
\end{align} 

\subsection{Action Loss}
\label{sec:action_loss}

The sequence loss ($L_{seq}$) only considers views within the same sequence, thus it does not incorporate similarity across sequences. We therefore introduce a second loss, which we term action loss, that measures the similarity of action segments across views. The loss is illustrated in Figure~\ref{fig:main_method_fig} (bottom).

For the action loss, we randomly sample an action segment $s_{i, \alpha}^{q} \in \mathbb{R}^{T_{s_{i}} \times H}$ from sequence $i$, where $\alpha$ denotes the action label of the segment and $T_{s_{i}}$ the frames of the segment. We uniformly sample a second action segment $s_{j, \alpha}^{r}$ with the same action label $\alpha$, but a different view $ r \neq q$. In this case, $j$ can be the same or another sequence.  

As in the sequence loss, the embeddings $z_{i, \alpha}^{q}$ and $z_{j, \alpha}^{r}$ are obtained by $\mathcal{E}_{\theta}$ and the predictor $\mathcal{P}_{\psi}$ computes $p_{i, \alpha}^{q}$. The cosine similarity between $p_{i, \alpha}^{q}$ and $z_{j, \alpha}^{r}$ is then given by 
\begin{align}
    \mathcal{S}(p_{i, \alpha}^{q}, z_{j, \alpha}^{r}) = \sum_{t=1}^{\min(T_{s_{i}},T_{s_{j}} )} \mathcal{D}(p_{i, \pi_i(t), \alpha}^{q}, z_{j, \pi_j(t), \alpha}^{r}) 
    \label{eq:action_simlarity}
\end{align}
where $\pi_i$ and $\pi_j$ are a temporal alignment between the segments $s_{i, \alpha}^{q}$ and $s_{j, \alpha}^{r}$. We use a linear alignment where $\pi(t)=t$ for the shortest segment and $\pi$ is a linear subsampling for the longest segment.     

Finally, the symmetric version of the action loss is given by 
\begin{align}
    \mathcal{L}_{action} = -\sum_{i} \frac{1}{2} \big[\mathcal{S}(p_{i, \alpha}^{q}, z_{j, \alpha}^{r}) + \mathcal{S}(p_{j, \alpha}^{r}, z_{i, \alpha}^{q}) \big].
    \label{eq:action_loss}
\end{align}
In the experiments, we show that both loss terms complement each other. While the action loss aims that the representation of actions across sequences and views is similar, the sequence loss makes a frame-by-frame comparison of different views of the same sequence. While the latter can also include frames that have not been annotated, the action loss is limited to annotated action segments.

\subsection{Joint Training}
\label{sec:joint_training}
The action segmentation model $\mathcal{E}_{\theta}$ and predictor head $\mathcal{P}_{\psi}$ are trained jointly using the following loss:
 \begin{align}
 \mathcal{L} = \mathcal{L}_{TAS} + \lambda \mathcal{L}_{seq} + \beta \mathcal{L}_{action}.
 \label{eq:total_loss}
\end{align}
$\mathcal{L}_{TAS}$ denotes the action segmentation loss from~\citep{farha2019ms}, which consists of the frame-wise cross entropy and a smoothing mean squared error (MSE) loss. The impact of the sequence loss and the action loss are steered by $\lambda$ and $\beta$, respectively, which are evaluated in the experiments. For $\mathcal{E}_{\theta}$, we use the publicly available network~\citep{ltc2023bahrami}.

\section{Experiments}
\label{sec:expr}
\textbf{Datasets.} 
Current datasets for action segmentation, such as Breakfast~\citep{breakfast} and 50Salads~\citep{50salads}, lack a multi-view setting. Specifically, 50Salads is captured using a single camera view, while the Breakfast dataset, recorded with 3 to 5 cameras, suffers from uncalibrated cameras and varying camera positions depending on the location~\citep{breakfast}. 
We evaluate the performance of our method on Assembly101~\citep{sener2022assembly101} and IkeaASM~\citep{BenShabat2020TheIA} since both datasets contain multiple views and are densely annotated by the occurring actions. In addition, we evaluate our approach on EgoExoLearn~\citep{huang2024egoexolearn}, which does not provide multiple views for training but different views for training and testing.

\textbf{Assembly101}~\citep{sener2022assembly101} is a procedural activity dataset featuring multi-view videos depicting people assembling and disassembling 101 toy vehicles. The dataset comprises 4,321 videos, amounting to a total of 513 hours, captured from 12 distinct viewpoints. Among these 12 views, 8 are exocentric (third-view), and the remaining 4 are egocentric (first-person view). The videos have a maximum duration of 25 minutes, with an average of 24 coarse-grained action segments per video. These segments are densely annotated, covering actions from 202 coarse-grained action classes consisting of 11 verbs and 61 objects. Moreover, the dataset consists of 1380 fine-grained action classes categorized by objects and interaction verbs.
Following the action segmentation benchmark of~\citep{sener2022assembly101}, we utilize the training split and evaluate the model on the validation set.

\textbf{IkeaASM}~\citep{BenShabat2020TheIA} is the multi-view dataset of IKEA furniture assembly. In the videos, 48 different participants assemble 4 furniture types (drawer, TV bench, coffee table, side table) with 3 different colors (oak, white, black) in 5 distinct environments (home, lab, office). In total, the dataset contains 1,113 videos with an overall duration of 35.16 hours. These videos were captured from three exocentric viewpoints. The videos are an average of 1.89 minutes long and contain segments from 33 atomic action classes. The dataset is split into training and testing sets with 762 and 351 videos, respectively. The dataset includes three views capturing the front, side, and top views of the work area.

\textbf{EgoExoLearn}~\citep{huang2024egoexolearn} is a dataset featuring demonstration videos paired with egocentric videos in which the camera wearers replicate the demonstrated tasks in a different environment. The dataset focuses on asynchronous procedural activities captured from both ego- and exo-views. EgoExoLearn comprises 747 video sequences totaling 120 hours of footage, covering activities from daily food preparation to laboratory experiments.  

\begin{figure*}[t!]
    \centering
    \includegraphics[width=\linewidth]{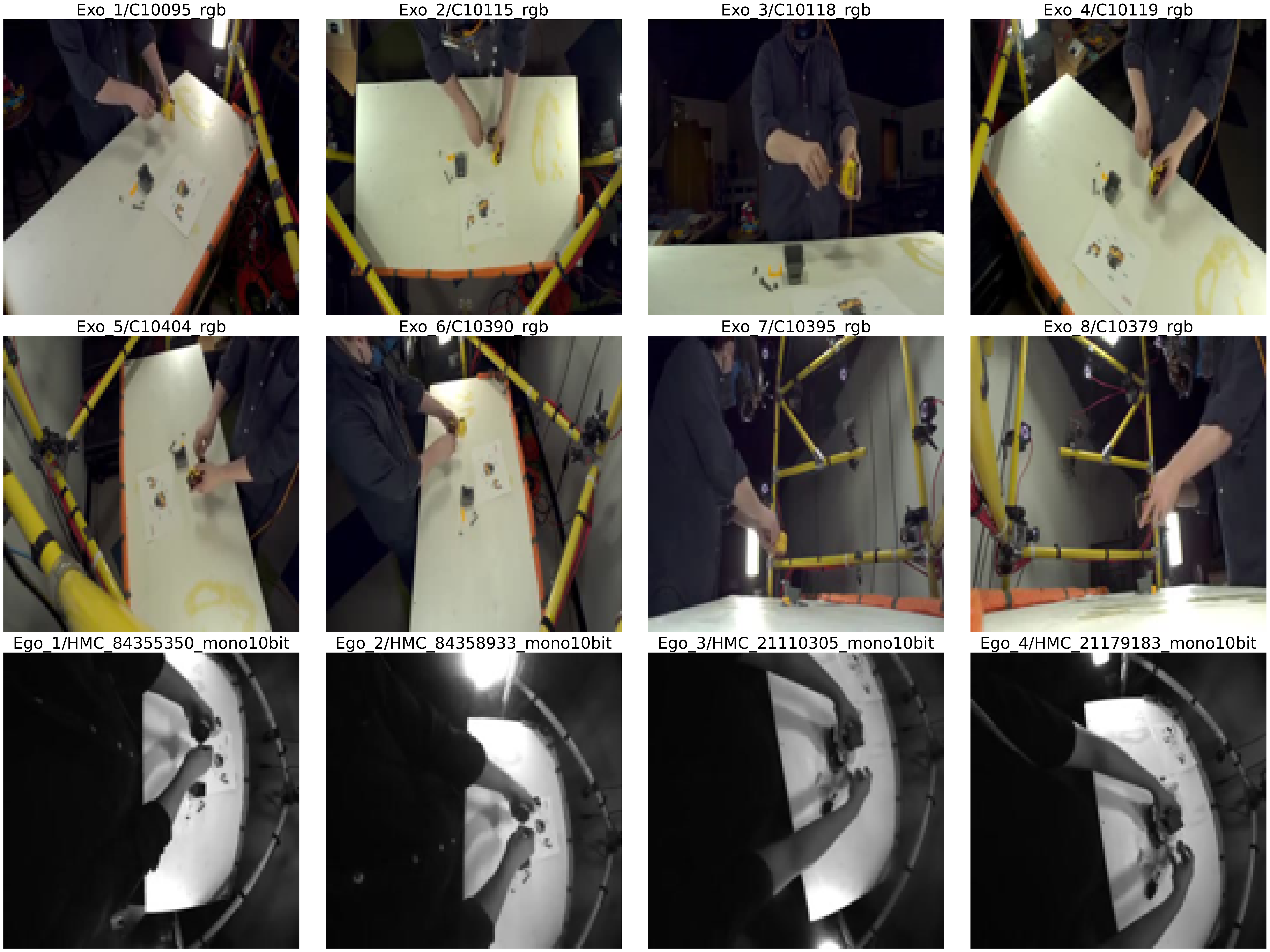}
    \caption{Camera views of the Assembly101 dataset. We use $\mathcal{V}_{seen} = \{Exo_{1}, \dots, Exo_{6}\}$ as seen views. As unseen views, we use $\mathcal{V}^{exo}_{unseen} = \{ Exo_{7}, Exo_{8} \}$ and $\mathcal{V}^{ego}_{unseen} = \{ Ego_{1}, Ego_{2}, Ego_{3}, Ego_{4} \}$. }
    \label{fig:assembly_views}
\end{figure*}

\textbf{Multi-View Evaluation Setting.}
In all datasets, we split the set of available views $\mathcal{V} = \mathcal{V}_{\text{seen}} \cup \mathcal{V}_{\text{unseen}}$ into $\mathcal{V}_{\text{seen}} \subset \mathcal{V}$ and  $\mathcal{V}_{\text{unseen}} = \mathcal{V} \setminus \mathcal{V}_{\text{seen}}$. The seen views are used for training, while we evaluate the model's performance on all views $\mathcal{V}$ with particular focus on $\mathcal{V}_{\text{unseen}}$.
As \textit{Assembly101} contains two distinct types of views, namely exocentric (exo) and egocentric (ego), we partition $\mathcal{V}_{\text{unseen}} = \mathcal{V}_{\text{unseen}}^{\text{exo}} \cup \mathcal{V}_{\text{unseen}}^{\text{ego}}$ to analyze the ego and exo views separately. Each sequence has 10 views, including 8 exocentric (exo) views and 2 egocentric (ego) views. Specifically, $\mathcal{V}_{\text{seen}}$ comprises 6 exo views, while both $\mathcal{V}_{\text{unseen}}^{\text{exo}}$ and $\mathcal{V}_{\text{unseen}}^{\text{ego}}$ contain two views per sequence.
Figure~\ref{fig:assembly_views} shows the views of the {Assembly101} dataset. For our protocol, we use $\mathcal{V}_{seen} = \{Exo_{1}, \dots, Exo_{6}\}$ as seen views. As unseen views, we use $\mathcal{V}^{exo}_{unseen} = \{ Exo_{7}, Exo_{8} \}$ and $\mathcal{V}^{ego}_{unseen} = \{ Ego_{1}, Ego_{2}, Ego_{3}, Ego_{4} \}$. Each sequence has 10 views with some recorded only from $Ego_{1}$ and $Ego_{2}$ and others from $Ego_{3}$ and $Ego_{4}$. For evaluation, we use all unseen views that are available.
For \textit{IkeaASM}, we train on two views and utilize one additional view as unseen view for testing.
For \textit{EgoExoLearn}, we follow the zero-shot Exo-Only setting~\citep{huang2024egoexolearn}.

For the input features, we use self-supervised DinoV2 features~\citep{oquab2023dinov2} without fine-tuning on Assembly101, IkeaASM, and EgoExoLearn. Additionally, we use MViTv2~\citep{li2022mvitv2},
initialized from masked feature prediction pre-training~\citep{wei2022masked} and fine-tuned exclusively using seen views on the Assembly101 dataset, following the action recognition protocol outlined in~\citep{sener2022assembly101}. 
Our protocol including the features and models will be made available upon acceptance of the paper.

\textbf{Evaluation Metrics.} We report the frame-wise accuracy (Acc), segmental Edit distance, and segmental F1 score at the overlap thresholds of $10\%$, $25\%$, and $50\%$ denoted by $F1 @ \{10, 25, 50 \}$. The F1 score uses the intersection over union (IoU) ratio for measuring overlap. For EgoExoLearn, we follow~\citep{huang2024egoexolearn} and report $F1@Avg$, which is the average of F1 over all thresholds $\{10, 25, 50 \}$. The edit score measures the order of actions without considering duration. Frame-wise accuracy measures accuracy per frame, with a bias towards actions of longer duration and less sensitivity to over-segmentation errors where a few frames within a ground-truth action segment are misclassified. The F1 score is considered the most reliable measure. For a more detailed description of the metrics, we refer to~\citep{ding2023tasSurvey}.

\subsection{Ablation Studies}\label{sec:ablation}
In the following, we evaluate the impact of hyperparameters and design choices and compare our approach to a Baseline model, which is the action segmentation model trained on seen views without additional loss terms or modifications. For action segmentation, we use the model proposed by~\citep{ltc2023bahrami}. We present all results using MViTv2 features on the Assembly101 dataset. The evaluation metrics are reported separately for Seen, Unseen Exo, and Unseen Ego views, as it is crucial to analyze the impact of Seen views in addition to Unseen views.

\begin{figure*}[tb]
    \centering
    \includegraphics[width=1.0\textwidth]{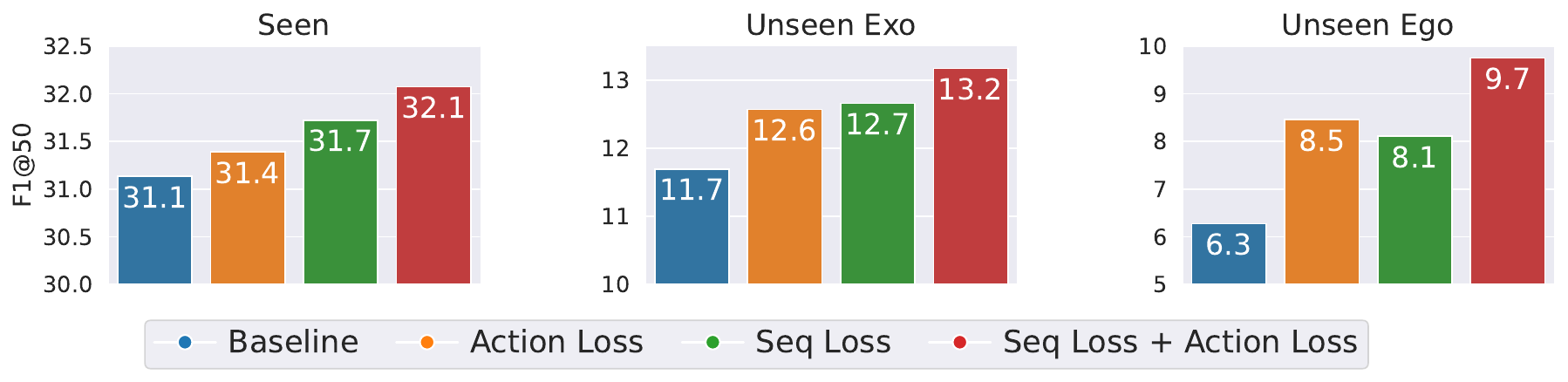}
    \caption{The F1@50 score shows the impact of adding the Sequence Loss ($\mathcal{L}_{seq}$) and Action Loss ($\mathcal{L}_{action}$) to the Baseline for MViTv2 features. Sequence Loss denotes the addition of $\mathcal{L}_{seq}$ to the Baseline, while Action Loss denotes adding $\mathcal{L}_{action}$ to the Baseline.}
    \label{plot:loss_ablations_mvit}
\end{figure*}

\begin{table}[t]
\caption{Impact of using different loss terms using MViTv2 features. $\mathcal{L}_{action}$ (s.\ view) denotes that the action loss also includes action segments of the same view.}
\label{tab:loss_ablation_mvit}
\centering
\setlength{\tabcolsep}{5.2pt}
\begin{tabular*}{\textwidth}{l cccc | cccc | cccc}
\toprule
\multirow{2}{*}{} & 
\multicolumn{4}{c|}{$\mathcal{V}_{seen}$}  &
\multicolumn{4}{c|}{$\mathcal{V}_{unseen}^{exo}$} & 
\multicolumn{4}{c}{$\mathcal{V}_{unseen}^{ego}$}  \\
\cmidrule{2-13}  
& \multicolumn{2}{c}{F1@\{25, 50\}} & Edit  & Acc 
& \multicolumn{2}{c}{F1@\{25, 50\}} & Edit  & Acc
& \multicolumn{2}{c}{F1@\{25, 50\}} & Edit  & Acc \\
\hline
Baseline  &
38.8 & 31.1  & 35.6 & 50.4 &
16.6 & 11.7 & 18.6 & 25.6 &  
9.4 & 6.3 & 11.4 & 17.2    \\

\hline
+ $\mathcal{L}_{seq}$ &
39.0 & 31.7 & 35.7 & 50.9 &
18.2 & 12.7 & \textbf{19.7} & \textbf{29.9} &  
12.4 & 8.1 &  15.2 & 22.0    \\
 \hline
 + $\mathcal{L}_{action}$ &
39.0 & 31.4 & 35.8 & 51.0 &
17.7 & 12.6 & 19.7 & 27.8 &  
13.2 & 8.5 &  15.8 & 23.2    \\
 \hline
+ $\mathcal{L}_{seq}$ + $\mathcal{L}_{action}$  &
\textbf{39.6} &\textbf{ 32.1} & \textbf{36.3} &\textbf{ 51.4} &
\textbf{18.8} & \textbf{13.2}  & 19.4 & 28.4 &  
\textbf{15.2} & \textbf{9.7}  & \textbf{16.7} & \textbf{24.8} \\ 
 \hline
 + $\mathcal{L}_{seq}$ + $\mathcal{L}_{action}$ (s.\ view)  &
 38.4 & 30.7  & 35.0 & 50.1 &
    17.7 & 12.4 & 18.9 & 27.3 &  
    11.7 & 7.0 & 14.5 & 20.2    \\
    \hline
+ $\mathcal{L}_{seq}$ + $\mathcal{L}_{action}$ + $\mathcal{L}_{Adv}$  &
38.8 & 31.0 & 35.6 & 50.2 &
17.6 & 12.5  & 18.6 & 28.3 &  
8.8 & 6.4  & 11.4 & 17.2 \\ 
\bottomrule
\end{tabular*}
\end{table}


\textbf{Impact of Different Loss Terms.} Figure~\ref{plot:loss_ablations_mvit} shows the F1@50 scores when using the action-level loss $\mathcal{L}_{action}$, the sequence-level loss $\mathcal{L}_{seq}$, or their combination. We first compare the two variants, using only $\mathcal{L}_{seq}$ or $\mathcal{L}_{action}$. For both unseen ego and exo views, adding either the action loss or sequence loss leads to improvements while the best results are achieved by combining both loss terms. 
In particular, the F1@50 improvement relative to the Baseline for unseen ego and exo views is 54\% and 12.8\%, respectively. The improvement in unseen ego views is more pronounced than in unseen exo views. This difference can be attributed to the substantial dissimilarity between ego and exo views, leading to a greater benefit from using the two loss terms for ego views.
Furthermore, the results for seen views also show improvement, potentially due to the regularization effect of the additional loss terms.  
Furthermore, we present the results for all metrics in Table~\ref{tab:loss_ablation_mvit}. While the edit score and accuracy are the highest for using only the sequence loss on Unseen Exo, the combination of loss terms performs best for all other settings. In particular, the F1@50 score is highest for the combination and our method outperforms the Baseline across all metrics. 
We also investigated the impact of computing the action-level loss using only action segments from different views, as described in Section~\ref{sec:action_loss}. If we compute the action-level loss also for action segments of the same view, which is denoted by $\mathcal{L}_{action}$ (s.\ view) in Table~\ref{tab:loss_ablation_mvit}, the performance drops since the action-level loss enforces then less the learning of a view-independent action representation.
Additionally, we evaluated the impact of adding an adversarial loss $\mathcal{L}_{Adv}$. To this end, we add a view classification head for predicting the view of each input to the model. We maximize a cross-entropy loss applied to the output of the view classification head, which can be interpreted as a discriminator as in Generative Adversarial Networks~\citep{goodfellow2014generative}. Maximizing the loss of the view classification head aims to eliminate view-dependent information, making the model more robust to unseen views. Table~\ref{tab:loss_ablation_mvit} shows that incorporating an adversarial loss, alongside the action-level loss $\mathcal{L}_{action}$ and the sequence-level loss $\mathcal{L}_{seq}$, leads to a performance drop in both unseen ego and exo views. In Section~\ref{sec:comparisons}, we also compare our approach to a setup where the adversarial loss is added to the Baseline.

\begin{figure}[tb]
    \centering
     \includegraphics[width=1.0\linewidth]{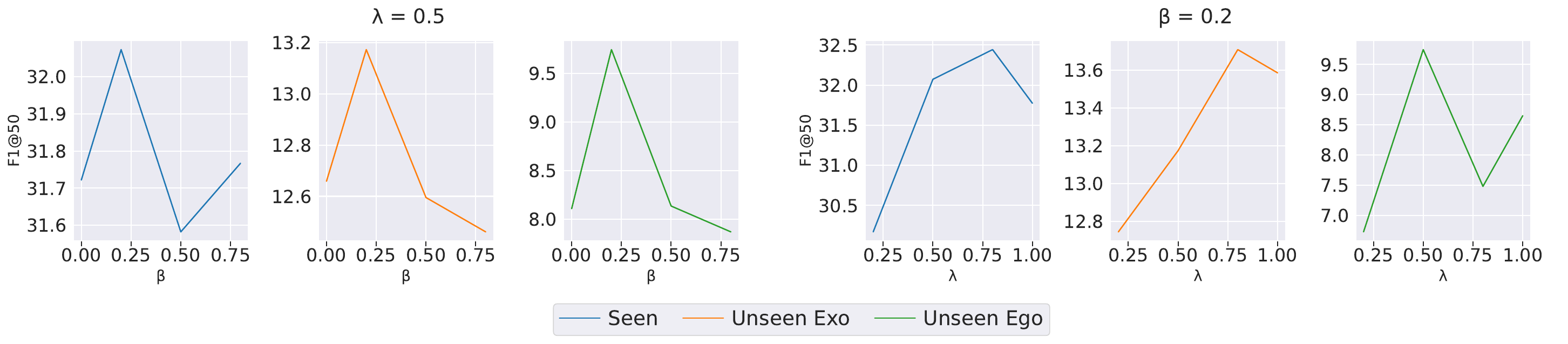}
    \caption{The plots show the influence of weighting between $\mathcal{L}_{seq}$ and $\mathcal{L}_{action}$, where $\lambda$ represents the weight of $\mathcal{L}_{seq}$ and $\beta$ represents the weight of $\mathcal{L}_{action}$. In the left plots, $\lambda$ is fixed at $0.5$ while $\beta$ varies. In the right plots, $\beta$ is set at $0.2$ while $\lambda$ varies.}
    \label{fig:loss_weight_barplot}
\end{figure}

\textbf{Impact of Loss Weights.}
Figure~\ref{fig:loss_weight_barplot} presents the performance of using different weights for the sequence loss $\mathcal{L}_{seq}$ and the action loss $\mathcal{L}_{action}$. Here, $\lambda$ represents the weight for the sequence loss, while $\beta$ denotes the weight for the action loss. As the figure suggests, the optimal values for $\lambda$ and $\beta$ are 0.5 and 0.2, respectively, considering both seen and unseen views.
Fixing $\lambda$ to 0.5 and increasing $\beta$ results in a performance drop for seen and unseen ego and exo.
On the other hand, as depicted in the right plot of Figure~\ref{fig:loss_weight_barplot}, increasing $\lambda$ while fixing $\beta$ to 0.2 does not improve the F1@50. It slightly increases for unseen exo, but slightly decreases for unseen ego.

\begin{figure}[tb]
    \centering
     \includegraphics[width=1.0\linewidth]{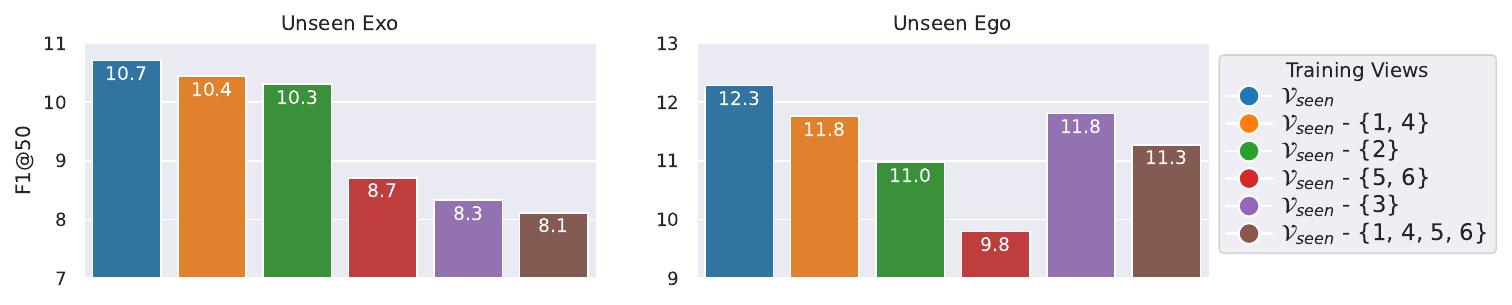}
    \caption{The plots show the influence of using different views for training. The numbers $\{1, 2, 3, 4, 5, 6\}$ refer to the ID of Exo view (Figure~\ref{fig:assembly_views}). For example, view with id 3 is $Exo_3$. }
    \label{fig:abl_views}
\end{figure}

\textbf{Impact of Training Views.} 
Figure~\ref{fig:abl_views} shows the impact of using different views for training. For our model, we use all views from $\mathcal{V}_{\text{seen}}$. 
In this experiment, we analyze how removing certain views from the training set affects the performance of our model on both seen and unseen views. We group views 5 and 6 together, as well as views 1 and 4, due to their visual symmetry (see Figure~\ref{fig:assembly_views}). Additionally, we consider a scenario where four views—namely views 1, 4, 5, and 6—are removed from the training set. This grouping is based on the similarity between views 1 and 4, as well as views 5 and 6, which could potentially reduce the impact of the reduced number of training samples to some extent. Training with all six views achieves highest overall performance across all unseen views. Interestingly, removing views 5 and 6 leads to a dramatic drop in the F1@50 score for unseen Ego views, likely due to the visual similarity between views 5 and 6 and the Ego views. Removing four views results in a 2.6-point reduction in the F1@50 score for unseen Exo views compared to using all views. However, the reduction for Ego views is only 1.0, which is less than the 2.5-point reduction observed when removing views 5 and 6 alone.

\begin{figure}[tb]
    \centering
     \includegraphics[width=\textwidth]{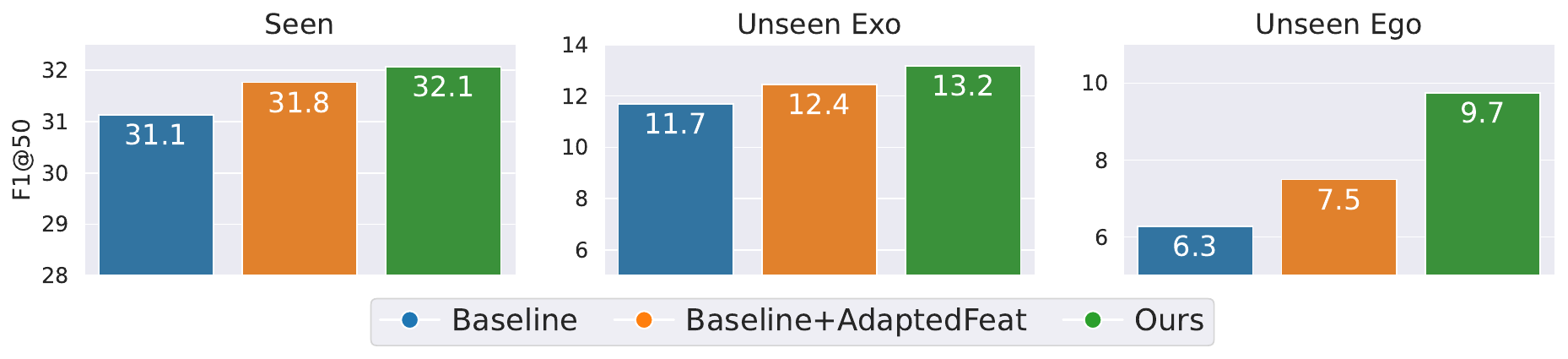}
    \caption{
    Applying the action loss to MViTv2 features (AdaptedFeat) and then training an action segmentation method is less effective than applying our loss to the action segmentation method directly (Ours). 
    }
    \label{fig:adapted_feat_barplot}
\end{figure}

\begin{figure}[tb]
    \centering
     \includegraphics[width=\textwidth]{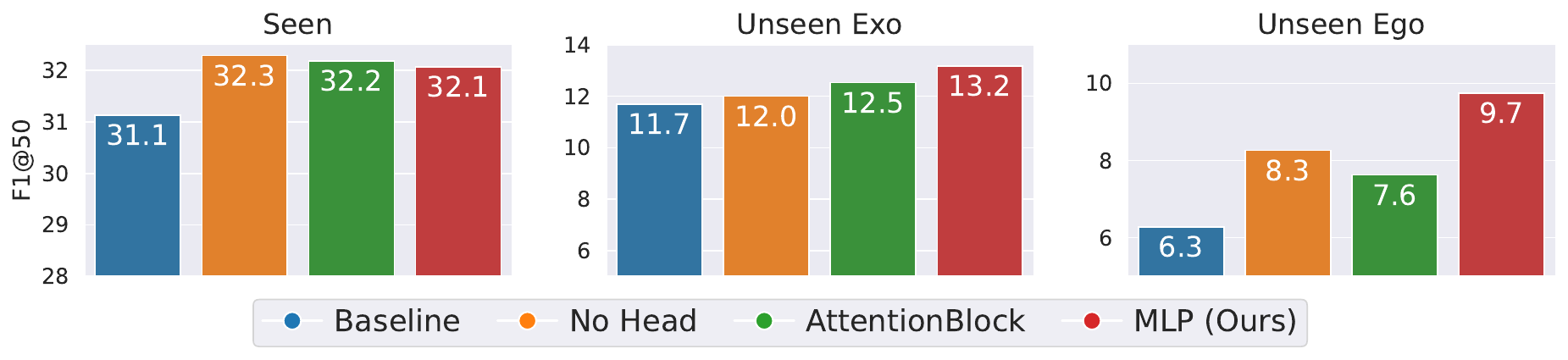}
    \caption{F1@50 for using different modules for the predictor head. For the AttentionBlock, we use the block from~\citep{ltc2023bahrami}.}
    \label{fig:predictor_head_abl}
\end{figure}

\textbf{Applying the Action Loss On Adapted Features.}
To evaluate the influence of input features on our method, we design an experiment to adapt the features of the pre-trained MViTv2. We freeze the existing layers of MViTv2 and introduce an additional Transformer encoder layer~\citep{vaswani2017attention} on top. The features are then adapted using the action loss $\mathcal{L}_{action}$, which we call AdaptedFeat. Then we train our baseline method with these features as inputs. Figure~\ref{fig:adapted_feat_barplot} shows the result of this experiment in terms of F1@50. While AdaptedFeat shows improvements over the Baseline, it falls short of achieving the results obtained by our method, highlighting the importance of learning the generalization across views at the level of action segmentation instead of feature level.

\textbf{Impact of Predictor Head.}
For the predictor head $\mathcal{P}_{\psi}$, we use a 3-layer MLP with non-linearity. In Figure~\ref{fig:predictor_head_abl}, we compare the results of ablating the predictor head. We consider two scenarios: one where we entirely remove the predictor head, and another where we replace the MLP with an AttentionBlock. For the AttentionBlock, we use the block from~\citep{ltc2023bahrami}. 
While the performance without the predictor head shows a slight improvement of 0.2 over our method on seen views, using MLP achieves the highest F1@50 for all unseen views.

\textbf{Impact of Fine and Coarse Action Segments.} 
Assembly101 offers annotations at two levels of granularity, namely fine-grained and coarse-grained. Table~\ref{tab:coarse_vs_fine_segm_abl} shows the results of using fine or coarse action segments for the action loss. It is important to note that the action segmentation task utilizes the coarse segments of Assembly101~\citep{sener2022assembly101}. While the utilization of coarse action segments is superior to fine segments in unseen exo by 0.4 in terms of F1@50, using action segments outperforms coarse segments by 3.2 on unseen ego. Furthermore, seen views also benefit from the fine segments more than from the coarse segments.

\begin{table}[tb]
\caption{Impact of using the coarse segments instead of fine segments in the action loss ($\mathcal{L}_{action}$).}
\label{tab:coarse_vs_fine_segm_abl}
\centering
\setlength{\tabcolsep}{1.0pt}
\begin{tabular}{l cccc | cccc | cccc}
\toprule
\multirow{2}{*}{Segment Granularity} & 
\multicolumn{4}{c|}{$\mathcal{V}_{\text{seen}}$}  &
\multicolumn{4}{c|}{$\mathcal{V}_{unseen}^{exo}$} & 
\multicolumn{4}{c}{$\mathcal{V}_{unseen}^{ego}$}  \\
\cmidrule{2-13} 
& \multicolumn{2}{c}{F1@\{25, 50\}} & Edit  & Acc 
& \multicolumn{2}{c}{F1@\{25, 50\}} & Edit  & Acc
& \multicolumn{2}{c}{F1@\{25, 50\}} & Edit  & Acc \\
\hline
Coarse  &
38.5 & 30.5  & 35.0 & 50.5 &
18.2 & \textbf{13.6} & 19.1 & 27.9 &  
10.0 & 6.5 & 12.6 & 18.8    \\
\hline
Fine  &
\textbf{39.6} &\textbf{ 32.1} & \textbf{36.3} &\textbf{ 51.4} &
\textbf{18.8} & 13.2  & \textbf{19.4} & \textbf{28.4} &  
\textbf{15.2} & \textbf{9.7}  & \textbf{16.7} & \textbf{24.8} \\  
\bottomrule
\end{tabular}
\end{table}

\begin{figure*}[tb]
    \centering
    \includegraphics[width=1.0\textwidth]{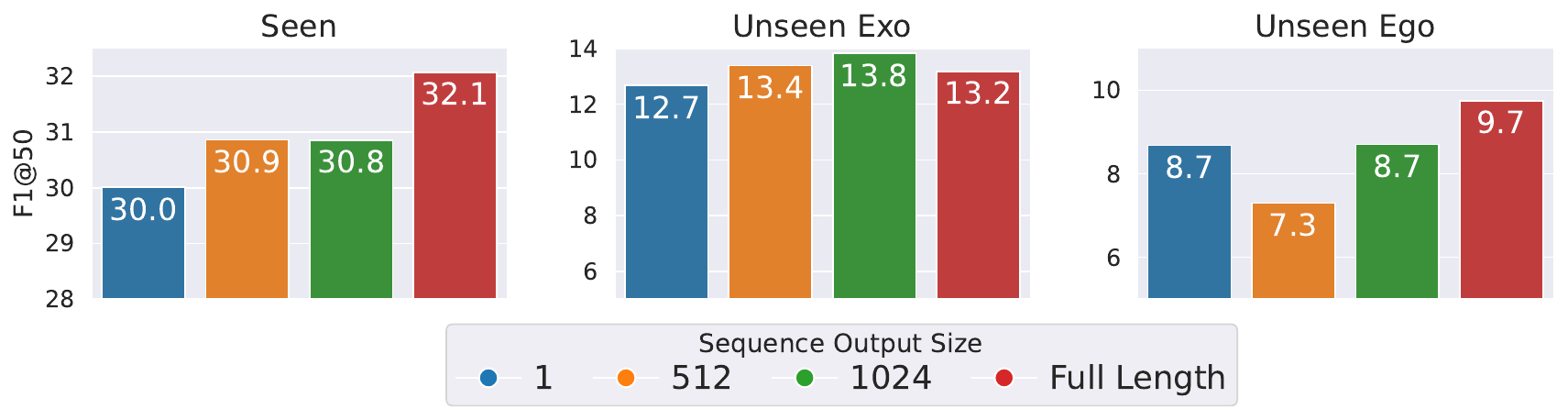}
    \caption{The F1@50 score of pooling the output sequences to 1, 512, 1024 and using the full length without any pooling in the sequence loss ($\mathcal{L}_{seq}$) using MViT features.}
    \label{plot:seq_pool_ablations_mvit}
\end{figure*}

\textbf{Impact of Pooling Sequences and Action Segments.}
We do not apply any pooling over the sequences in the sequence loss or the action segments in the action loss. Here, we present the result of pooling the sequences and action segments before computing the cosine similarity. 
Figure~\ref{plot:seq_pool_ablations_mvit} illustrates the impact of pooling the sequences to a length of 1, 512, and 1024. Pooling the sequences results in a decrease of F1@50 for unseen ego views. For unseen exo views, however, it can lead to a small improvement of 0.6. In general, when considering both seen and unseen views, pooling the sequences decreases F1@50 in average. 

\begin{figure*}[tb]
    \centering
    \includegraphics[width=1.0\textwidth]{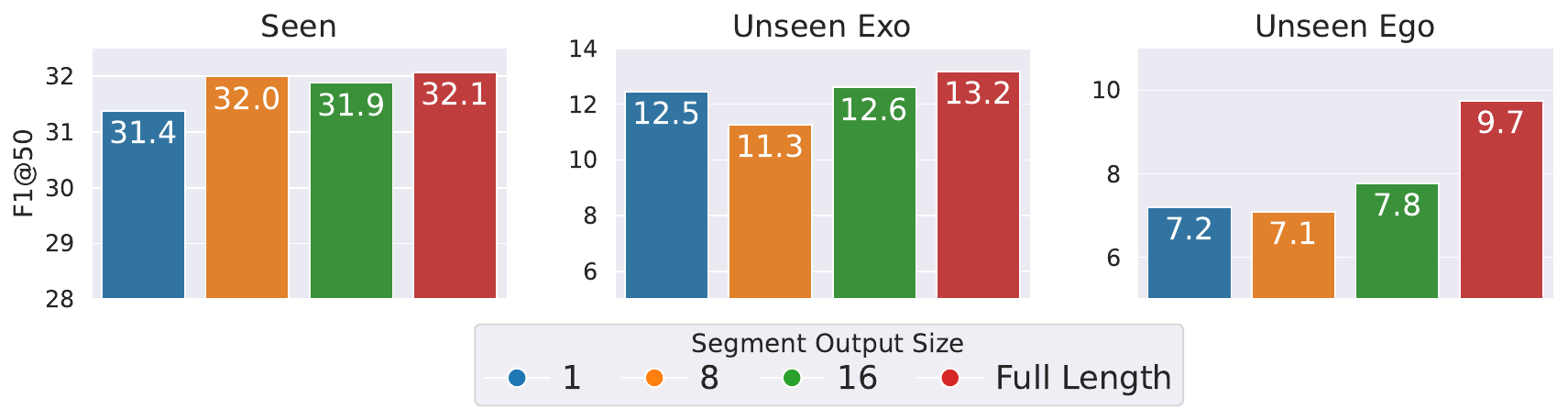}
    \caption{ The F1@50 score for pooling the output segments to 1, 8, 16 and keeping full length without any pooling in the action loss ($\mathcal{L}_{action}$) using MViT features.}
    \label{plot:segment_pool_ablations_mvit}
\end{figure*}

Figure~\ref{plot:segment_pool_ablations_mvit} shows the result of pooling action segments. In this case, the F1@50 scores decrease from 9.7 to 7.2 for unseen ego when pooling to 1.

\textbf{Impact of Cosine Similarity.}
To evaluate the impact of the cosine similarity, we replace the cosine similarity with the Mean Squared Error (MSE) and the Kullback–Leibler (KL) divergence. In this case, we replace \eqref{eq:seq_loss} and \eqref{eq:action_loss} by MSE or KL divergence, respectively. 
Table~\ref{tab:cosine_vs_mse_kl_abl} shows the results of this experiment in terms of F1@50. The cosine similarity outperforms MSE across all seen and unseen views. While the KL divergence performs on par with the cosine similarity for seen views, it performs worse for unseen views.

\begin{table}[tb]
\caption{Impact of replacing the cosine similarity with Mean Squared Error (MSE) and Kullback–Leibler (KL) divergence loss.}
\label{tab:cosine_vs_mse_kl_abl}
\centering
    \setlength{\tabcolsep}{3.5pt}
    \begin{tabular}{l cccc | cccc | cccc}
    \toprule
    \multirow{2}{*}{} & 
    \multicolumn{4}{c|}{$\mathcal{V}_{\text{seen}}$}  &
    \multicolumn{4}{c|}{$\mathcal{V}_{unseen}^{exo}$} & 
    \multicolumn{4}{c}{$\mathcal{V}_{unseen}^{ego}$}  \\
    \cmidrule{2-13} 
    & \multicolumn{2}{c}{F1@\{25, 50\}} & Edit  & Acc 
    & \multicolumn{2}{c}{F1@\{25, 50\}} & Edit  & Acc
    & \multicolumn{2}{c}{F1@\{25, 50\}} & Edit  & Acc \\
    \hline
    MSE  &
    37.5 & 30.0 & 34.4 & 49.9 &
    17.2 & 11.9 & 18.4 & 27.6 &  
    10.6 & 7.2 & 11.6 & 18.0    \\
    \hline
    KL divergence  &
    39.4 & \textbf{32.4} & 36.1 & \textbf{51.6} &
    17.9 & 12.3 & 18.5 & 27.5 &  
    9.6 & 6.5 & 11.8 & 19.4    \\
    \hline
    Cosine  &
    \textbf{39.6} & 32.1 & \textbf{36.3} & 51.4 &
    \textbf{18.8} & \textbf{13.2}  & \textbf{19.4} & \textbf{28.4} &  
    \textbf{15.2} & \textbf{9.7}  & \textbf{16.7} & \textbf{24.8} \\  
    \bottomrule
    \end{tabular}%
 
\end{table}

\begin{figure*}[tb]
    \centering
    \includegraphics[width=1.0\textwidth]{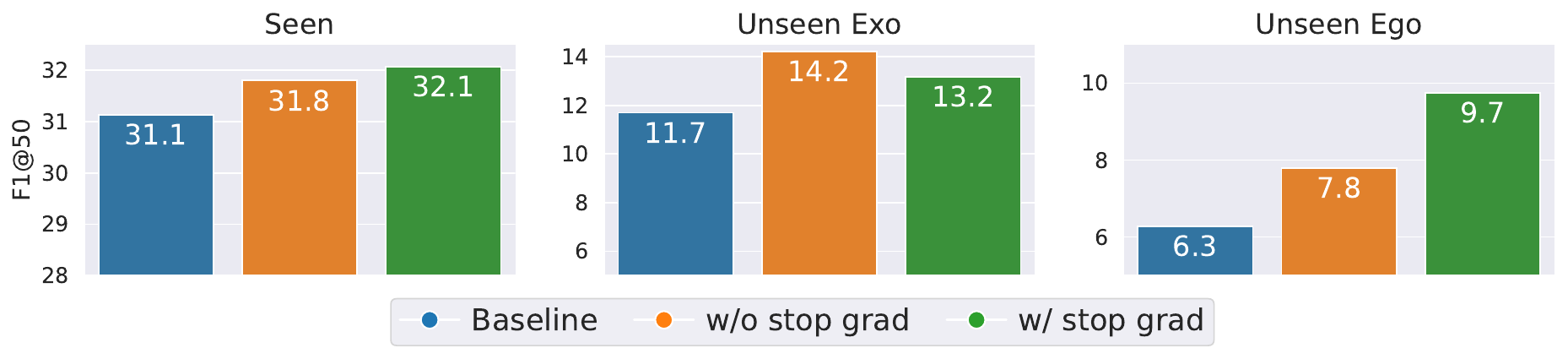}
    \caption{Ablation of using stopping gradient.}
    \label{plot:stop_grad_abl}
\end{figure*}

\textbf{Using Stopping Gradient.}
Figure~\ref{plot:stop_grad_abl} evaluates the effect of the stopping gradient in our model. 
While the work \citep{chen2021simsiam} suggests that stopping gradient is necessary to prevent collapse in self-supervised training, we do not observe such an effect here. Nevertheless, stopping gradient has a positive impact on seen and unseen ego, but results in a decrease in F1@50 for unseen exo.

\begin{table}[tb]
\caption{Impact of shifting the frames by a random offset in the range of $ [0, \delta ]$.}
\label{tab:frame_shift_expr}
\centering
    \setlength{\tabcolsep}{3.5pt}
    \begin{tabular}{l cccc | cccc | cccc}
    \toprule
    \multirow{2}{*}{} & 
    \multicolumn{4}{c|}{$\mathcal{V}_{\text{seen}}$}  &
    \multicolumn{4}{c|}{$\mathcal{V}_{unseen}^{exo}$} & 
    \multicolumn{4}{c}{$\mathcal{V}_{unseen}^{ego}$}  \\
    \cmidrule{2-13} 
    $\delta$ & \multicolumn{2}{c}{F1@\{25, 50\}} & Edit  & Acc 
    & \multicolumn{2}{c}{F1@\{25, 50\}} & Edit  & Acc
    & \multicolumn{2}{c}{F1@\{25, 50\}} & Edit  & Acc \\
    \hline
    Baseline  &
38.8 & 31.1  & 35.6 & 50.4 &
16.6 & 11.7 & 18.6 & 25.6 &  
9.4 & 6.3 & 11.4 & 17.2    \\
    \hline
    0  &
    39.6 & 32.1 & 36.3 & 51.4 &
    18.8 & 13.2  & 19.4 & 28.4 &  
    15.2 & 9.7  & 16.7 & 24.8 \\  
    1  &
    39.7 & 32.1 & 36.2 & 52.2 &
    17.3 & 12.3 & 18.6 & 27.2 &  
    11.5 & 7.7 & 13.9 & 20.0    \\
    10  &
    39.1 & 31.7 & 35.6 & 50.7 &
    17.6 & 12.8 & 18.6 & 26.5 &  
    8.9 & 6.3 & 10.6 & 15.7    \\

    \bottomrule
    \end{tabular}%
 
\end{table}

\textbf{Impact of View Synchronization.}
For the sequence loss, we assume that the camera views of the training data are synchronized. Table~\ref{tab:frame_shift_expr} evaluates the impact of the synchronization accuracy. A random shift of one frame between the views does not impact the performance for seen views, but it reduces the performance on unseen views. Nevertheless, it is still better than the Baseline. A random shift by 10 frames, does not decrease the performance on unseen exocentric views further, but it leads to a major drop in performance on unseen egocentric views. In this case, the annotations and frames are not any more aligned between the views, making it difficult to separate changes due to different views from changes due to time delays. This shows that the synchronization of the camera views is very important for achieving a generalization to unseen views.             

    \begin{table}[tb]
    \caption{
    Results on Assembly101 using ResNet3D, MViTv2, and  DinoV2 features. We compare to the baseline method~\citep{ltc2023bahrami} and DVANet~\citep{siddiqui2024dvanet}.
    }
    \label{tab:main_result_assembly}
    \centering
    \setlength{\tabcolsep}{3.5pt}
    {\footnotesize %
    \begin{tabularx}{\linewidth}{p{1.2cm}  l ccccc | ccccc | ccccc }
    \toprule 
    \multirow{2}{*}{feature}  &
    \multirow{2}{*}{method}   & 
    \multicolumn{5}{c|}{$\mathcal{V}_{\text{seen}}$}  &
    \multicolumn{5}{c|}{$\mathcal{V}_{unseen}^{exo}$} &
    \multicolumn{5}{c}{$\mathcal{V}_{unseen}^{ego}$} \\
    \cmidrule{3-17} 
    &   & 
    \multicolumn{3}{c}{F1@\{10, 25, 50\}} & Edit  & Acc & 
    \multicolumn{3}{c}{F1@\{10, 25, 50\}} & Edit & Acc  &
    \multicolumn{3}{c}{F1@\{10, 25, 50\}} & Edit & Acc \\ 
    \cmidrule{1-17} 
    \multirow{2}{*}{ResNet3D} & DVANet  &
    36.5 & 33.1 & 25.9 & 31.5 & 44.2 &
    18.0 & 15.7 & 10.2 & 17.8 & 22.4 &  
    5.6 & 4.2 & 2.3 & 7.9 & 6.8    \\
    \cmidrule{3-17} 
    & Ours  &
    \textbf{37.3} & \textbf{34.1} & \textbf{27.0} & \textbf{32.2} & \textbf{45.9} &
    \textbf{18.5} & \textbf{16.3} & \textbf{11.1 } & \textbf{18.0} & \textbf{23.4} &  
    \textbf{6.5} & \textbf{4.9} & \textbf{2.8 } & \textbf{9.1} & \textbf{7.6} \\  
    \midrule
    \midrule
     \multirow{3}{*}{MViTv2L} & 
    Baseline  &
     41.7 & 38.4 & 31.1  & 35.6  &  50.4 &
    19.7  & 16.6 & 11.7 & 18.6 &  26.5 &   
    10.8 & 9.4  & 6.3 & 11.4 & 17.2   \\
    \cmidrule{3-17} 
    & AdvLoss  &
     42.2 & 38.9 & 30.9 & 35.3 &  50.7 &
     20.5  & 17.8 & 12.6 & 19.0 & 27.2 &   
    14.8 & 12.4 & 8.3 & 14.7 & 20.9   \\
    \cmidrule{3-17}
    & Contrastive  &
     42.5 & 39.3 & 31.7 & 36.0 &  \textbf{51.8} &
     20.2  & 16.9 & 12.1 & 18.8 & 27.6 &   
    13.8 & 11.9 & 8.8 & 13.9 & 21.3   \\
    \cmidrule{3-17} 
    &  Ours &
    \textbf{42.6 } & \textbf{ 39.6} & \textbf{32.1} & \textbf{36.3}  & 51.4 &
    \textbf{21.3} & \textbf{18.8} & \textbf{13.2} & \textbf{19.4} & \textbf{ 28.4} & 
    \textbf{17.4} & \textbf{15.2} & \textbf{9.7} & \textbf{16.7} & \textbf{24.8} \\
    \midrule
    \midrule
    \multirow{3}{*}{Dinov2L}  & 
    Baseline  &
    33.8  &  29.5  & 21.3  & 29.9 &  40.8  &
    19.8  &  15.9   & 10.0    & 19.2  &  22.9   &   
    21.2  &  17.2   & 10.4   &  20.0  &  24.4   \\
    \cmidrule{3-17} 
    & AdvLoss  &
     \textbf{36.1} & \textbf{31.1} & \textbf{22.3} & \textbf{ 32.1} & \textbf{41.9} &
     19.1 & 15.6 & 10.2 & 19.5 & 22.5 &   
    \textbf{ 21.7} & 18.1 & 11.6 & \textbf{21.0} & 25.3   \\
    \cmidrule{3-17}
    & Contrastive  &
     34.2 & 30.4 & 22.1 & 30.9 &  41.7 &
     19.8  & 16.8 & \textbf{10.9} & 19.6 & \textbf{24.8} &   
    19.8 & 17.0 & 10.8 & 18.8 & 24.0   \\
     \cmidrule{3-17} 
     & Ours &
    34.4 &  29.9 & 21.7  & 30.5  & 41.4  &
    \textbf{20.8} & \textbf{17.2} & {10.7} &  \textbf{20.1 } & {23.7}  & 
    20.9 & \textbf{18.2} & \textbf{12.3} &  20.6 & \textbf{25.5} \\
    \bottomrule
    \end{tabularx}%
    }
    \end{table}

\subsection{Comparisons}\label{sec:comparisons}

\textbf{Assembly101.}
In Table~\ref{tab:main_result_assembly}, we present the results of our method on Assembly101 using ResNet3D, DinoV2, and MViTv2 features. We compare our method with a Baseline~\citep{ltc2023bahrami} where the action segmentation model is trained on seen views without the loss terms that are introduced in this work. We also compare our approach with two alternative approaches. AdvLoss extends the Baseline model with an adversarial loss, which is described in Section~\ref{sec:ablation}. The loss aims to eliminate view-dependent information, making the model more robust to unseen views. Contrastive extends the Baseline model with a contrastive loss~\citep{oord2018representation} that is computed using positive pairs formed by the same sequences from different views within a batch.

Furthermore, we compare our approach to a recent method for view-invariant feature learning, namely DVANet~\citep{siddiqui2024dvanet}. We utilized the public code of DVANet, which uses ResNet3D as backbone, and fine-tuned the model on the Assembly101 dataset. The learned view-invariant features are then used as input features for temporal action segmentation~\citep{ltc2023bahrami}. 

The results in Table~\ref{tab:main_result_assembly} show that our approach outperforms action segmentation based on DVANet~\citep{siddiqui2024dvanet} in all settings. Using ResNet3D as backbone and DVANet~\citep{siddiqui2024dvanet} for feature learning performs poorly when there is a very large difference between the views in the training data and the views in the test data ($\mathcal{V}_{unseen}^{ego}$). When we compare our approach to the Baseline~\citep{ltc2023bahrami} and AdvLoss in Table~\ref{tab:main_result_assembly}, we observe that our method consistently exhibits improvements over both Baseline and AdvLoss across all metrics using MViTv2 features. For instance, on unseen ego views, the F1@50, Edit, and accuracy (Acc) metrics see improvements of $+3.4$, $+5.3$, and $+7.6$ over the Baseline, respectively. Furthermore, the performance on seen views is also improving, potentially due to the regularization impact of the introduced loss terms. 

While the Contrastive method achieves competitive results on seen views using MViTv2 features, it falls short on both unseen exocentric and egocentric views, showing a 9\% reduction in F1@50 on unseen ego.

\begin{figure}[t]
    \centering
    \includegraphics[width=1.0\textwidth]{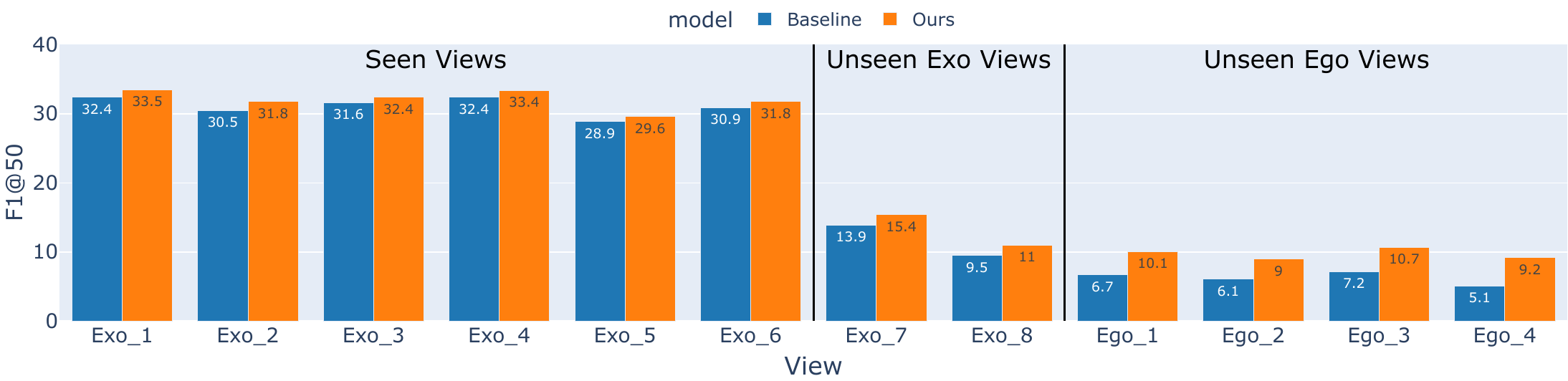}
    \caption{The F1@50 score per view for the Assembly101 dataset using MViT features.}
    \label{fig:per_view_mvit_f1}
\end{figure}

\begin{figure}[tb]
    \centering
    \includegraphics[width=1.0\textwidth]{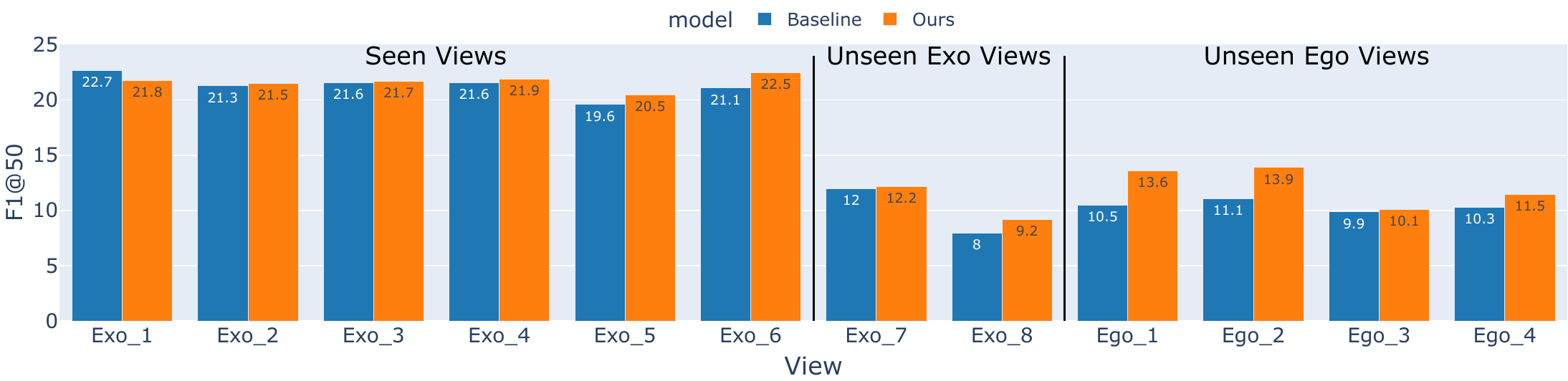}
    \caption{The F1@50 score per view for the Assembly101 dataset using Dino features.}
    \label{fig:per_view_dino_f1}
\end{figure}

For DinoV2 features, our method outperforms the Baseline, AdvLoss, and Contrastive for most metrics on unseen views as well. While the Contrastive method achieves slightly higher accuracy on unseen exocentric views, AdvLoss achieves a slightly higher Edit distance on unseen egocentric views.
Figures~\ref{fig:per_view_mvit_f1} and \ref{fig:per_view_dino_f1} show the F1@50 score per view for Assembly101 using MViTv2 and Dino features. Our model consistently outperforms the Baseline across all views, with particularly significant improvements for unseen ego views.

\begin{figure}[t]
    \centering
    \begin{subfigure}{0.48\textwidth}
        \centering
        \includegraphics[width=\textwidth]{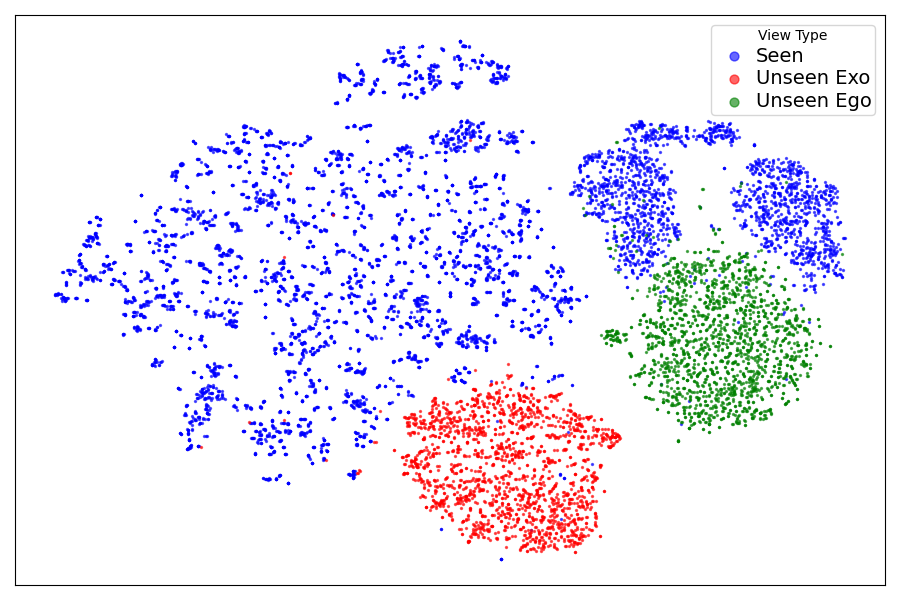}
        \caption{}
        \label{fig:mvit_input_feat}
    \end{subfigure}
    \hfill
    \begin{subfigure}{0.48\textwidth}
        \centering
        \includegraphics[width=\textwidth]{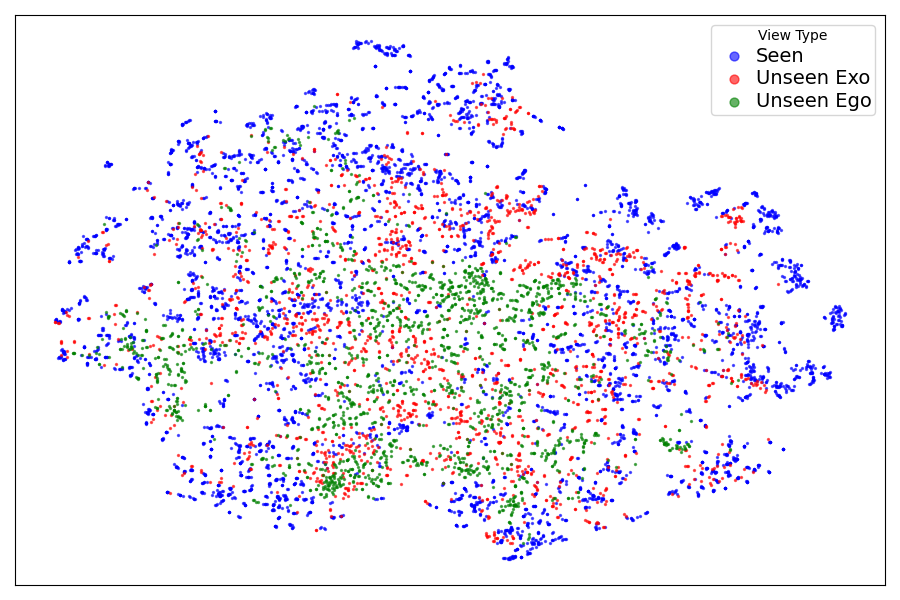}
        \caption{}
        \label{fig:ours_mvit_feat}
    \end{subfigure}
    \caption{t-SNE visualization comparing MViTv2 features (left) with features learned by our model (right).  
    }
    \label{fig:mvit_feat_tsne}
\end{figure}

To visualize how our approach discards view information, Figure~\ref{fig:mvit_feat_tsne} shows t-SNE visualizations of the MViTv2 features and the learned features of our segmentation model. The three colors in Figure~\ref{fig:mvit_input_feat} represent seen, unseen exocentric, and unseen egocentric views, forming distinct and well-separated clusters. This separation indicates that the MViTv2 features are view-dependent. In contrast, Figure~\ref{fig:ours_mvit_feat} shows that the seen and unseen views are not separated anymore, suggesting increased similarity in feature representations across views.

\textbf{IkeaASM.}
Table~\ref{tab:main_result_ikea} shows the results on the IkeaASM dataset using DinoV2 features. In all metrics on unseen views, our method outperforms the Baseline, AdvLoss, and Contrastive, except for the Edit score, which is very close to the Baseline. 

Specifically, it achieves the highest $F1@50$ scores for both seen and unseen views, with values of 48.0 for seen views and 47.7 for unseen views. Additionally, our method has the highest Accuracy (70.0) for unseen views.

\begin{table}[tb]
\caption{
Results on IkeaASM using DinoV2.
}
\label{tab:main_result_ikea}
\centering
\setlength{\tabcolsep}{3.5pt}
\begin{tabular}{l  ccccc | ccccc }
\toprule
\multirow{2}{*}{method}  & \multicolumn{5}{c|}{$\mathcal{V}_{\text{seen}}$}    & \multicolumn{5}{c}{$\mathcal{V}_{\text{unseen}}$} \\
\cmidrule{2-11} 
                         &  \multicolumn{3}{c}{F1@\{10, 25, 50\}} & Edit  & Acc &  \multicolumn{3}{c}{F1@\{10, 25, 50\}} & Edit & Acc \\ 
\hline 
Baseline  &
 69.4  & 63.9 & 47.1 & \textbf{63.8}  &  71.1 &
67.2 & 61.1  & 44.2 & \textbf{62.7} & 68.7  \\
\cmidrule{2-11}
AdvLoss  &
 67.7  & 61.8 & 47.5 & 61.7  &  69.2 &
65.5 & 60.2  & 45.1 & 60.7 & 68.1  \\
\cmidrule{2-11}
Contrastive  &
 69.5  & \textbf{64.2} & \textbf{48.7} & 63.1  &  \textbf{71.2} &
67.8 & 62.5  & 46.4 & 61.8 & 69.6  \\
\cmidrule{2-11}
Ours &
\textbf{69.7}  & 63.7 & 48.0 & 63.3 & 71.0 &
\textbf{68.4} & \textbf{62.9} & \textbf{47.7} & 62.6 & \textbf{70.0} \\
\bottomrule
\end{tabular}
\end{table}

\textbf{EgoExoLearn.}
The results for EgoExoLearn using I3D and DinoV2 features are presented in Table~\ref{tab:main_result_egoExoLearn}. Following~\citet{huang2024egoexolearn}, we report the F1@Avg, which is the average of F1@$\{10, 25, 50\}$. For EgoExoLearn, we use the zero-shot exo-only setting and report results on both the validation and test sets. Since the dataset contains only one training view per sequence, we utilize only the action loss for our approach.
Our method outperforms the method presented in \citep{huang2024egoexolearn} using the provided I3D features~\citep{huang2024egoexolearn} on both test and validation sets in terms of F1@Avg for both seen exo and unseen ego views. In addition to I3D features, we extract DinoV2 features and train the method of~\citep{huang2024egoexolearn} using these features. There is a noticeable improvement in overall action segmentation results when using DinoV2 features compared to I3D features.
For DinoV2 features, our method outperforms \citep{huang2024egoexolearn} for both test and validation sets on all metrics.

\begin{table}[tb]
\caption{
Results on the validation and test set of EgoExoLearn using I3D and DinoV2 features following the zero-shot setting of~\citep{huang2024egoexolearn} for action segmentation. * Indicates reported numbers from~\citep{huang2024egoexolearn}.}
\label{tab:main_result_egoExoLearn}
{\footnotesize%
\setlength{\tabcolsep}{1.0pt}
\begin{tabular}{l l c cc | ccc | c cc | ccc  }
\toprule
\multirow{3}{*}{feature} & \multirow{3}{*}{method}  & \multicolumn{6}{c|}{Val}  & \multicolumn{6}{c}{Test} \\
\cmidrule{3-14} 
& 
& \multicolumn{3}{c|}{Ego}    & \multicolumn{3}{c |}{Exo} 
& \multicolumn{3}{c|}{Ego}    & \multicolumn{3}{c}{Exo} \\
\cmidrule{3-14} 
& 
&  F1@Avg & Edit  & Acc 
&  F1@Avg & Edit  & Acc 
&  F1@Avg & Edit  & Acc 
&  F1@Avg & Edit  & Acc  \\ 
\midrule 
\multirow{3}{*}{I3D} & EgoExoLearn* &
 7.0  & \textbf{33.8}  &  \textbf{28.0} &
 23.6  &  40.3  &  \textbf{38.6}  &
 8.1  & 35.3  &  \textbf{24.8} &
 20.1  &  37.3  &  \textbf{42.6}  \\
\cmidrule{2-14}
 & Ours &
 \textbf{11.0}  & 33.5 & 26.2 &
 \textbf{26.8} & \textbf{41.1} &  36.5  &
 \textbf{9.8} & \textbf{35.7}  &  18.8 &
 \textbf{22.7} & \textbf{37.6}  &  36.6  \\
 \midrule
 \midrule
 \multirow{3}{*}{DinoV2\_L} & EgoExoLearn &
15.3  &  36.1  &  35.5  &
33.2  & 38.1  &  52.7 &
 11.7 & 33.2  & 21.4 &
 29.9 & 38.2 & 48.4  \\
\cmidrule{2-14}
& Ours &
\textbf{17.0} &\textbf{ 36.6} & \textbf{36.8} &
\textbf{32.5}  & \textbf{42.2} & \textbf{53.5} &
\textbf{14.9} & \textbf{34.3} & \textbf{32.5} &
\textbf{32.1}  & \textbf{38.4} & \textbf{51.4}  \\
\bottomrule
\end{tabular}
}
\end{table}

\begin{figure*}[tb]
    \centering
    \includegraphics[width=1.0\linewidth]{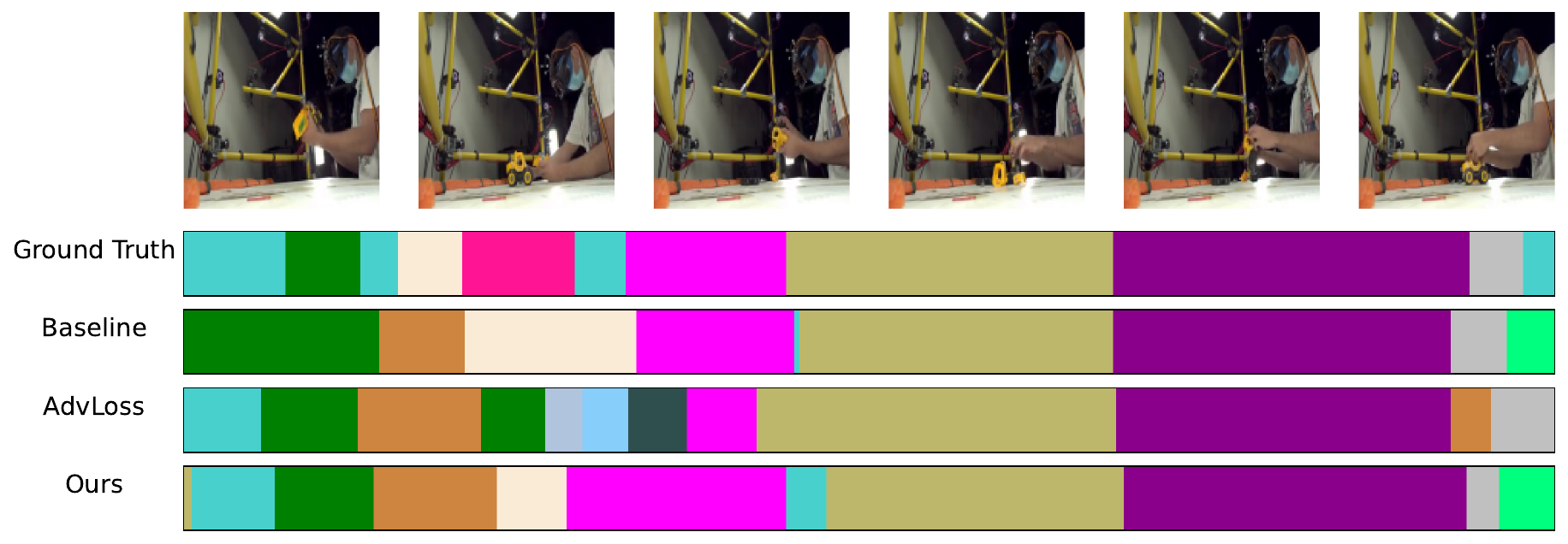}
    \caption{Qualitative results on an unseen exo view of Assembly101. }
    \label{fig:qual_assmbely_exo}
\end{figure*}

\begin{figure*}[tb]
    \centering
    \includegraphics[width=1.0\linewidth]{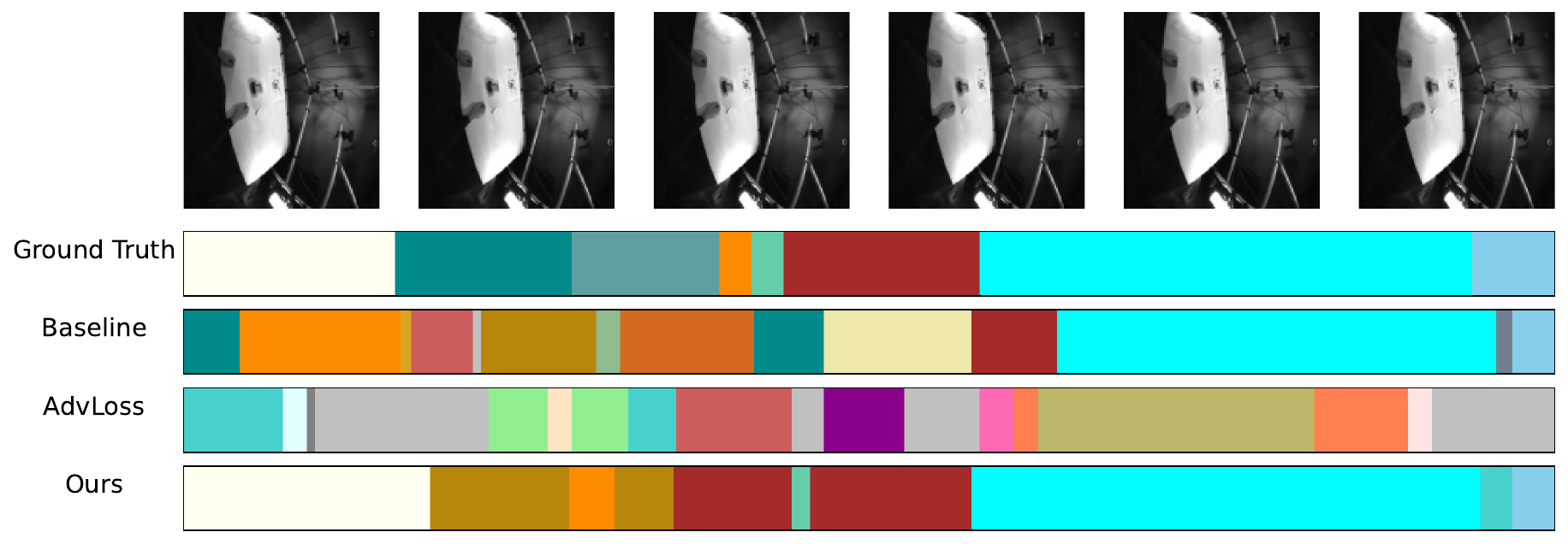}
    \caption{Qualitative results on an unseen ego view of Assembly101. }
    \label{fig:qual_assmbely_ego}
\end{figure*}

\begin{figure*}[tb]
    \centering
    \includegraphics[width=1.0\linewidth]{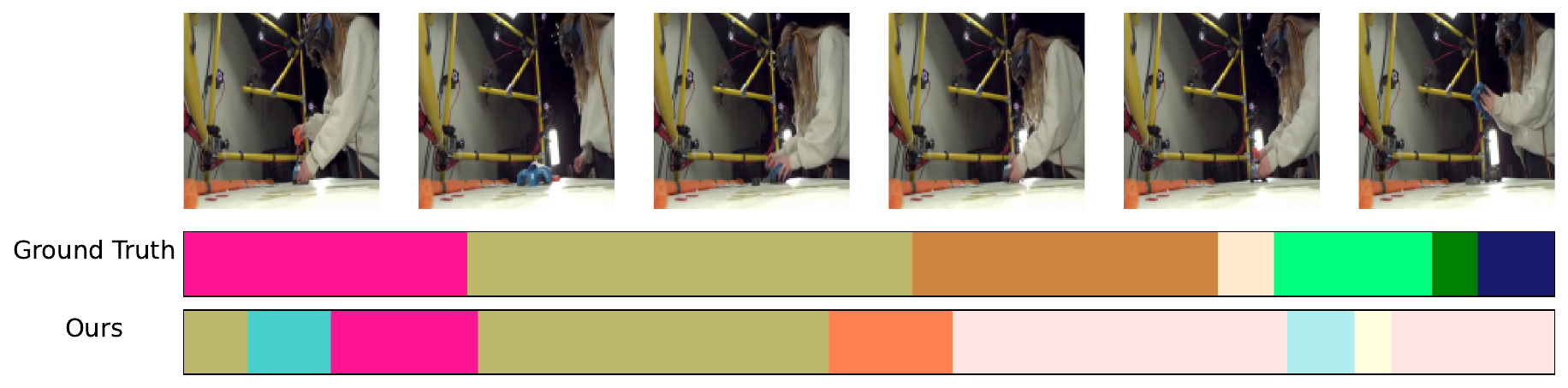}
    \caption{Failure case of our method on an unseen exo view of Assembly101. }
    \label{fig:failure_qual_assmbely_exo}
\end{figure*}

\begin{figure*}[tb]
    \centering
    \includegraphics[width=1.0\linewidth]{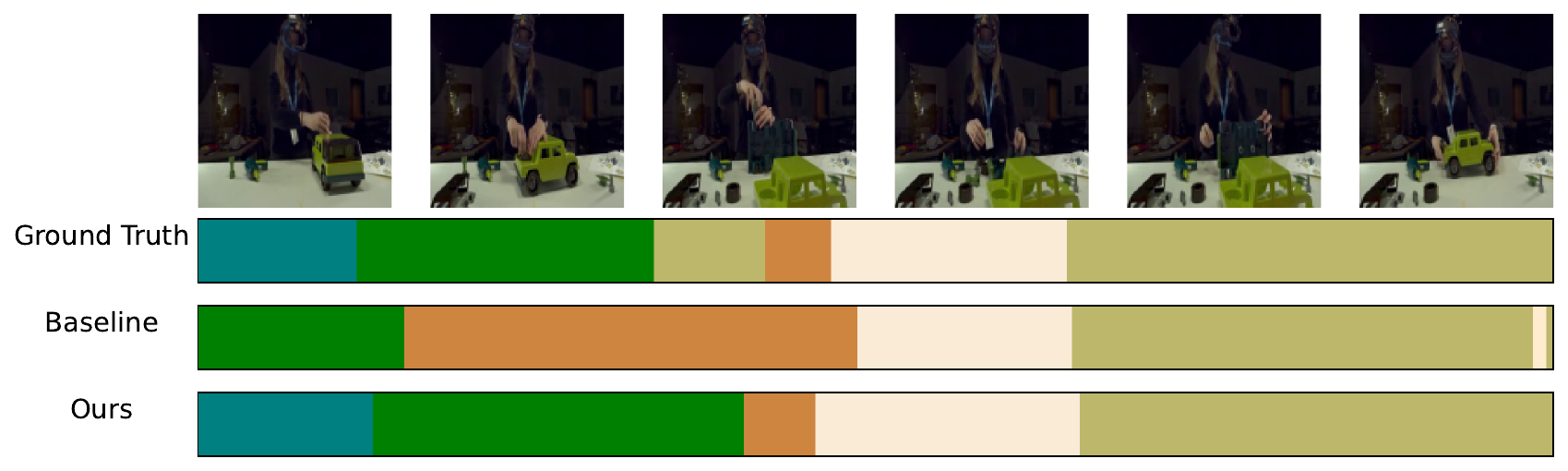}
    \caption{Qualitative results on an seen view of Assembly101. }
    \label{fig:qual_assmbely_seen}
\end{figure*}

\begin{figure*}[tb]
    \centering
    \includegraphics[width=1.0\linewidth]{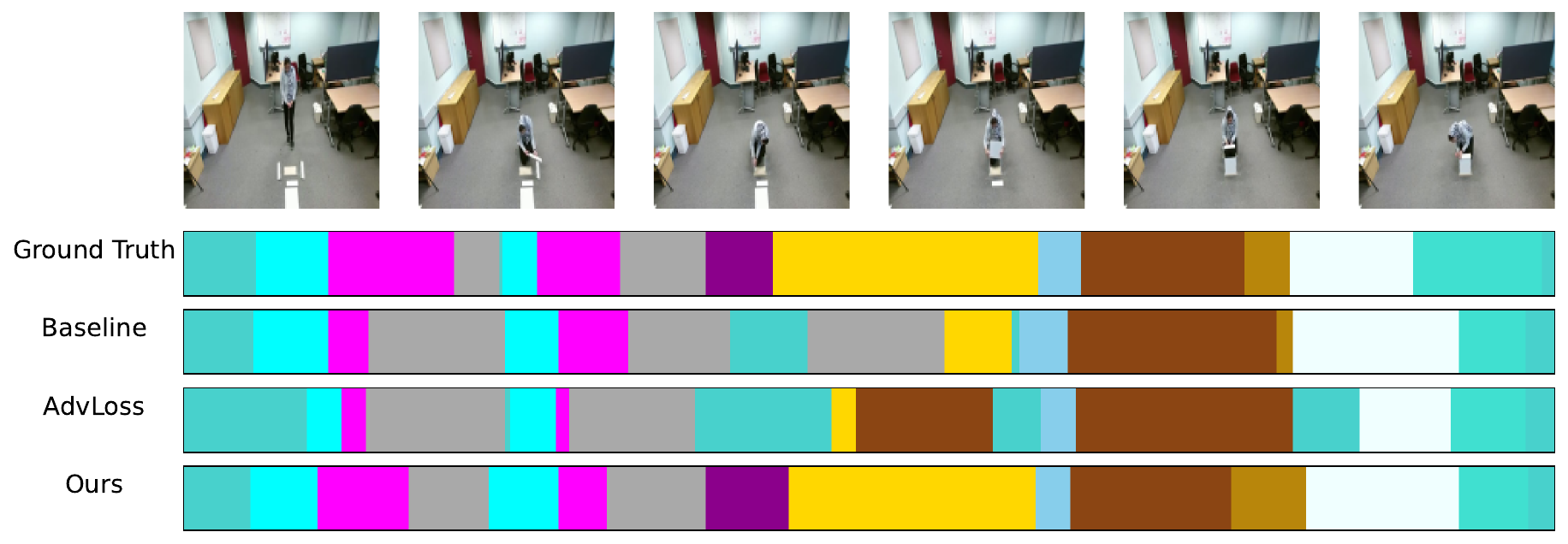}
    \caption{Qualitative results on an unseen view of the IkeaASM dataset. }
    \label{fig:qual_ikea}
\end{figure*}

\subsection{Qualitative Results.}
In Figures~\ref{fig:qual_assmbely_exo} and \ref{fig:qual_assmbely_ego}, we present some qualitative results for unseen views of Assembly101. Compared to the Baseline, our method provides more accurate predictions at the beginning of the video, where the Baseline fails to identify the blue action segment corresponding to \textit{inspect toy}.  While the predictions of AdvLoss are more accurate at the beginning of the video, it later identifies several action segments that do not occur in the video. For the unseen ego view shown in Figure~\ref{fig:qual_assmbely_ego}, the results are less accurate. Nevertheless, the predictions of our method are clearly better for the second half of the sequence compared to the Baseline and AdvLoss. 
Figure~\ref{fig:failure_qual_assmbely_exo} illustrates a failure case of our method on an unseen exo view from Assembly101. Our model struggles to predict the correct action segments toward the end of the video, where multiple actions are missed. For instance, the green-labeled ground truth actions at the end of the video, \textit{detach interior} and \textit{detach roof}, are incorrectly predicted as \textit{detach dump bed} and \textit{detach water tank}. Additionally, the dark blue action, \textit{inspect body}, is completely missed by our model.

Additionally, we present qualitative results on an seen view of Assembly101 in Figure~\ref{fig:qual_assmbely_seen}. Our predictions show less errors compared to the Baseline, however, it fails to capture an action segment corresponding to the \textit{detach wheel} action in the middle of the video.
Figure~\ref{fig:qual_ikea} presents qualitative results for the IkeaASM dataset. Our approach exhibits fewer errors compared to both the Baseline and AdvLoss. For instance, the action segment with the purple color corresponding to the action \textit{pick up bottom panel} is accurately predicted by our approach and missed in the predictions of other methods. Furthermore, the predictions of our approach demonstrate better quality in terms of the order and length of predicted action segments. 

\section{Conclusion}
\label{sec:conclusion}
In this work, we have addressed the problem of temporal action segmentation on unseen views by leveraging a Siamese network with weight sharing at two different temporal levels. 

Our approach aims to learn similarity across views in the video representation through a sequence loss and similarity across views in the action representation through an action loss. 
We have evaluated our approach on unseen views that have not been part of the training data on the Assembly101, IkeaASM, and EgoExoLearn datasets. Our approach outperformed the baselines as well as two very recent approaches \citep{siddiqui2024dvanet} and \citep{huang2024egoexolearn}, and demonstrated better generalization across views than previous works.    

Despite our approach's promising improvements, several limitations persist. Although our method can also be applied to single-view training data as in EgoExoLearn, it performs best if it has access to multi-view training data, which may not be feasible in all scenarios. While we achieved a 54\% improvement in F1@50 for unseen egocentric views, the generalization from exocentric to egocentric views remains a major challenge. Finally, we do not address changes in sensor modality, e.g., thermal or depth cameras, whereas a mix of RGB and grayscale cameras does not impose any restrictions.          
These limitations suggest that more research will be needed
to improve the generalization of temporal action
segmentation approaches to unseen views.

\backmatter

\bibliography{main}

\end{document}